\documentclass[lettersize,journal]{IEEEtran}
\usepackage{amsmath,amsfonts}
\usepackage[linesnumbered,ruled,vlined]{algorithm2e}
\usepackage{array}
\usepackage[caption=false,font=normalsize,labelfont=sf,textfont=sf]{subfig}
\usepackage{textcomp}
\usepackage{stfloats}
\usepackage{url}
\usepackage{amssymb}
\usepackage{verbatim}
\usepackage{amsmath}
\usepackage{graphicx}
\usepackage{cite}
\usepackage{orcidlink}
\PassOptionsToPackage{bookmarks=false}{hyperref} 
\hypersetup{hidelinks} 
\usepackage{optidef}

\newcommand{\hl}[1]{#1}

\begin{document}

\title{Reward-Based Collision-Free Algorithm for Trajectory Planning of Autonomous Robots}

\author{Jose D. Hoyos\orcidlink{0000-0001-8768-115X},~\IEEEmembership{Graduate Student Member,~IEEE}, Tianyu Zhou\orcidlink{0000-0003-2469-5911}, Zehui Lu\orcidlink{0000-0002-7312-7075}, and Shaoshuai Mou\orcidlink{0000-0002-3698-4238},~\IEEEmembership{Member,~IEEE}.


\thanks{The authors are with the School of Aeronautics and Astronautics, Purdue University, West Lafayette, IN 47907, USA (e-mail: mous@purdue.edu).}
\thanks{The supplementary video can be found at \url{https://youtu.be/iQ9f3bVbYis}.}
}



\maketitle

\begin{abstract}
\hl{This paper proposes a novel mission planning algorithm for autonomous robots that selects an optimal waypoint sequence from a predefined set to maximize total reward while satisfying obstacle avoidance, state, input, derivative, mission time, and distance constraints. The formulation extends the prize-collecting traveling salesman problem. A tailored genetic algorithm evolves candidate solutions using a fitness function, crossover, and mutation, with constraint enforcement via a penalty method. Differential flatness and clothoid curves are employed to penalize infeasible trajectories efficiently, while the Euler spiral method ensures curvature-continuous trajectories with bounded curvature, enhancing dynamic feasibility and mitigating oscillations typical of minimum-jerk and snap parameterizations. Due to the discrete variable length optimization space, crossover is performed using a dynamic time-warping-based method and extended convex combination with projection. The algorithm’s performance is validated through simulations and experiments with a ground vehicle, quadrotor, and quadruped, supported by benchmarking and time-complexity analysis.}
\end{abstract}

\def\abstractname{Note to Practitioners}
\begin{abstract}
\hl{This work addresses the challenge of managing multiple tasks within limited resources in cyber-physical systems, where tasks vary in priority and maximizing total reward is critical. We propose an evolutionary-inspired algorithm that optimizes task selection and prioritization in a motion planning context, generating trajectories that respect robot capabilities while favoring higher-priority tasks and potentially bypassing lower-priority ones to fulfill constraints. The algorithm integrates with traditional motion planning techniques, ensuring safe, resource-aware trajectories. It has been validated on a ground vehicle, a quadruped robot, and a quadrotor drone, demonstrating adaptability across platforms with varying constraints. Practical contributions include the method’s flexibility to accommodate different resource limitations, such as time, distance, and energy. The current formulation assumes static rewards and constraints; future work will explore real-time adaptation and extensions to broader classes of robots, including collaborative scenarios involving human-robot interaction.}
\end{abstract}

\begin{IEEEkeywords}
Motion-planning, Genetic algorithms, Mobile robot dynamics, Mobile robot kinematics
\end{IEEEkeywords}

\section{Introduction}

In recent decades, research in robotics has increasingly focused on enhancing autonomy capabilities, such as for multi-robot missions \cite{robotics13030040,PRASAD2022110391} and \hl{high-level human–robot cooperation} \cite{9849419,9852712,10639942}.
In this context, achieving autonomous operations requires motion planning strategies combined with feasible control inputs to guide the robot along the planned trajectory. This process must account for obstacle avoidance and adhere to dynamic and control input constraints, ensuring safe navigation in various environments, and so on.

\begin{figure}[!t]
\centering
\includegraphics[width=3in]{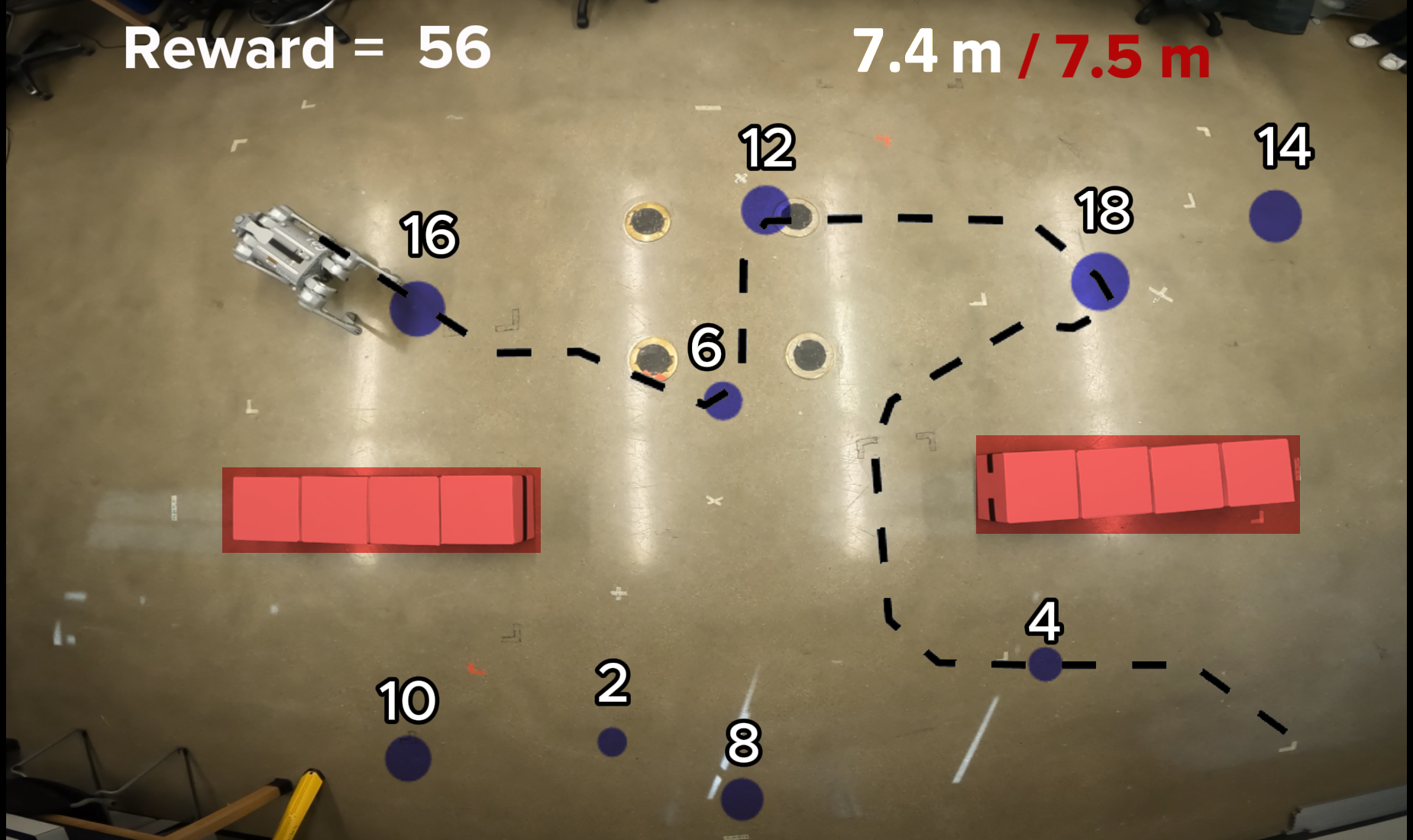}
\caption{Quadruped robot trajectory through weighted waypoints without a specific predetermined order, considering collision avoidance and total distance constraints. Larger blue dots represent higher rewards, while red boxes indicate obstacles. The trajectory maximizes the collected reward by prioritizing higher-reward waypoints, skipping lower-weighted ones, while satisfying the distance constraint.}
\label{fig:introduc}
\end{figure}

Existing research has proposed algorithms for trajectory and control planning in static environments, from a starting configuration to a final configuration \cite{9523743}. Similar algorithms can be employed when there are intermediate \hl{tasks, as long as they have a predefined order,} since the problem remains a classical motion planning and control scenario.
But when multiple waypoints need to be reached to complete various tasks without a predetermined order, the robot faces the challenge of solving both the motion planning problem and a combinatorial optimization problem. This situation commonly arises in scenarios such as warehouse automation, where robots pick items from diverse locations without a predefined sequence; agricultural robotics, where drones survey and treat different field sections based on dynamic factors like pest outbreaks or crop health; and autonomous delivery systems, where robots efficiently deliver packages to multiple destinations while navigating complex urban environments.

Furthermore, in many real-world applications, the number of available tasks can be substantial, often exceeding the mission’s allowable time, which is typically constrained by factors such as battery capacity. In these situations, it becomes crucial to bypass certain waypoints while maximizing the overall collected reward, as each waypoint or task carries a specific reward value. Fig. 1 illustrates this scenario, where the robot selectively skips less rewarding waypoints to adhere to the distance constraint. Some scenarios includes: drone delivery systems, where a drone with a low battery may prioritize delivering high-priority packages to critical locations while bypassing lower-priority deliveries, search and rescue operations, where robots may need to prioritize assisting the most critically injured individuals first, bypassing less critical waypoints to maximize mission success. Similarly, autonomous underwater vehicles used for environmental monitoring might prioritize collecting data from high-impact areas, such as regions with known pollution or biodiversity hotspots, bypassing less critical waypoints to ensure the collection of the most valuable data before battery depletion. We approach this scenario as a two-layer problem, \hl{where the primary layer addresses waypoint sequencing as a combinatorial optimization problem}, while the secondary layer concentrates on trajectory planning. Accordingly, we review the literature related to these two areas of research.

\subsection{Literature Review: Combinatorial Optimization}

Solving combinatorial optimization problems poses a significant challenge not only because it often results in exponential growth in solution space exploration (NP-complete \cite{gerkey2004formal}), but also because the discrete design space prevents the use of traditional gradient-based techniques. To tackle combinatorial problems, alternative methods such as neural networks \cite{Peng2021GraphLF} and heuristic methods such as genetic algorithms (GA) \cite{7130169} have been employed.
Despite having no theoretical guarantee of optimality, heuristic algorithms often produce good enough solutions in practice \cite{Peng2021GraphLF} and use fewer parameters compared to neural networks \cite{9626724}. Genetic algorithms, in particular, have been proven to be theoretically and empirically robust search methods \cite{10.5555/534133}.

The mission planning problem addressed in this paper can be seen as a variant of the rooted (fixed start) orienteering problem, also known as the bank robber problem or the generalized Traveling Salesman Problem (TSP), with additional constraints involving a maximum mission time window, an obstacle-free path between nodes, and different rewards assigned to each node. This is closely related to the prize-collecting traveling salesman problem. Furthermore, dynamics are added to the traveler as a constraint, including any state and input and \hl{their} derivatives. Due to the obstacle-free path requirement, the problem can be seen as a bi-level optimization since it involves trajectory planning between a given sequence.

Specifically, given a set of weighted waypoints distributed along a known static environment, the aim is to find a sequence of waypoints and an obstacle-free trajectory between them that maximizes the collected reward subject to specific maximum time and distance values, accounting for energy limitations and mission requirements. The problem is also subject to robot dynamics and control saturation, meaning that even if there is a path between two waypoints, it may involve a non-reachable trajectory due to input saturation or robot dynamics limitations. \hl{While the algorithms in the literature} \cite{6069411, 7185453,8613017,9196516, debnath2024integrated} \hl{provide valuable contributions, they typically do not address weighted waypoints, bypassing waypoints to meet constraints, or integrating dynamics into the motion planning constraints.}

Given the constraints, the optimal solution may not collect rewards from all waypoints, leading to solution candidates of varying dimensions. This variability prevents the use of traditional heuristic algorithms, such as particle swarm optimization, or standard crossover techniques in genetic algorithms, as these operations are not well-defined for elements with unmatched dimensions. To determine the optimal waypoint sequence in a variable-dimension space while satisfying dynamic constraints, a genetic algorithm (GA) is proposed and validated. The overall idea of a GA is the evolution of the candidate solutions using evolutionary-inspired rules. Each candidate is evaluated using a fitness function that assigns \hl{lower} values to the best individuals in the population. Then, based on these values, the next generation emerge from a crossover operation among the previous generation. An extra mutation step is usually added to increase diversity and exploration. This evolution eventually guides the population to high (or low) regions in the design space, finding the global optimal or close in many cases.

In this work, we incorporate the penalty method into the fitness function to account for constraints. During penalty evaluation, the differential flatness property is leveraged to efficiently map each candidate trajectory to its corresponding states and inputs, thereby assessing constraint fulfillment. Moreover, a two-strategy crossover scheme is employed: a subset of the offspring is generated using a crossover that combines dynamic time warping with an extended convex combination and projection, while the remaining subset is produced via the random subsequence insertion method. A similar warping approach is described in \cite{ha2021variable}. A preliminary obstacle-free path between any two waypoints is calculated during the fitness evaluation using the classical $A^*$ algorithm. From the $A^*$ intermediate waypoints, a trajectory parameterization method based on clothoid curves is proposed and compared to the classical snap and jerk polynomial methods, showing smoother shapes and curvature adaptability. Further discussion regarding the trajectory generation is presented in Section \ref{trajectory}. Finally, to enhance exploration capabilities and prevent premature convergence, the mutation step can swap two waypoints within \hl{a} given candidate sequence. The proposal is tested in simulations and validated using a quadrotor, ground vehicle, and a quadruped. A quadruped differential flatness map is proposed, based on the kinematic control strategy.

\subsection{Literature Review: Trajectory Generation}
\label{sec:trajectory_lit}

Trajectory generation involves computing a feasible, collision-free path that connects a sequence of waypoints while satisfying time and state-input constraints. This requires parameterizing the path by time and ensuring that the resulting trajectory remains dynamically feasible. In this section, we review existing approaches and their limitations.

\subsubsection{Trajectory Generation with Dynamics Consideration}
A fundamental challenge in trajectory generation is ensuring compliance with robot dynamics, defined by $\dot{\boldsymbol{x}}(t) = \boldsymbol{f}(\boldsymbol{x}(t),\boldsymbol{u}(t))$. Many methods address this by estimating states and inputs for a given waypoint sequence. Differentially Flat (DF) systems \cite{doi:10.1137/S0363012995274027,FLIESS1992159} have been widely used, as they allow trajectory parameterization through flat outputs, significantly reducing the complexity of feasibility verification.
Unlike traditional optimal control methods, DF-based approaches avoid solving differential equations, relying instead on algebraic constraints \cite{Murray2010OptimizationBasedC}.


\subsubsection{Existing Approaches in Trajectory Generation}
Model predictive control \cite{7798766} is commonly used for trajectory generation but is computationally expensive. Due to this limitation, real-time trajectory generation methods for DF systems have been developed, leveraging smooth basis functions to parameterize trajectories. Most DF-based methods use smooth basis functions $\boldsymbol{\phi}_j:[0,t_f]\mapsto \mathbb{R}^p$ to represent the flat output as a linear combination:
\begin{equation}
    \boldsymbol{y}(t) = \sum_{j=1}^{M} \beta_j\boldsymbol{\phi}_j(t),
\end{equation}
where the number of basis functions must be sufficient to satisfy all constraints, which are typically nonlinear and non-convex. Various basis functions have been used, including polynomials \cite{articlenice}, Legendre polynomials \cite{inproceedings123}, and splines \cite{9765821,1657645}.
In simpler cases, constraints are imposed only at initial and final times, with cost functions typically minimizing jerk or snap \cite{5980409}. These formulations often lead to standard quadratic programming problems \cite{articlenice} or, when no quadratic cost is minimized, linear programming problems.

For paths passing through multiple intermediate waypoints, trajectory generation is often solved using sampling-based methods such as $A^*$ and rapidly-exploring random trees. These methods divide the path into segments, requiring constraints at segment junctions to maintain smooth derivatives. While effective for a small number of segments \cite{5980409}, these approaches become numerically ill-conditioned when dealing with high-order polynomials, many segments, or widely varying segment times \cite{articlenice}.

\subsubsection{Limitations in State-of-the-Art Approaches}
Despite significant progress, trajectory generation methods exhibit key limitations.
Many studies address obstacle avoidance only as a secondary constraint, making generalization across different environments difficult \cite{Murray2010OptimizationBasedC, inproceedings123}.
Approaches for handling obstacles include control point adjustments \cite{inproceedings123} and artificial potential fields \cite{doi:10.1177/027836499101000604}, which can be interpreted as a type of control barrier function \cite{singletary2020comparative}. While Monte Carlo and Brownian motion techniques have been proposed to mitigate local minima issues \cite{doi:10.1177/027836499101000604}, they do not guarantee global feasibility.

Another major challenge is handling state and input constraints, particularly in underactuated systems. Formally, this requires finding a feasible set in the flat output space \cite{doi:10.2514/2.4732}, which is generally non-convex due to nonlinear inequalities \cite{articlenice}. The difficulty of incorporating such constraints can be seen in the following general form:
\begin{equation}
\label{saturations}
\boldsymbol{\mathcal{G}}(\boldsymbol{x},\Dot{\boldsymbol{x}},\dots,\boldsymbol{x}^{(r)}, \boldsymbol{u},\Dot{\boldsymbol{u}},\dots,\boldsymbol{u}^{(v)}) \leq \boldsymbol{0}, \quad \forall t \in [0, t_f],
\end{equation}
where $\boldsymbol{\mathcal{G}}$ represents general state and input constraints \cite{9765821}. While waypoint constraints impose a finite set of conditions on basis function coefficients, state and input constraints must hold at all times, making feasibility difficult or impossible to ensure for a finite number of basis functions. A common workaround is discretizing the constraints over a large set of time samples, as done for linear systems in \cite{AGRAWAL19991154}. To enforce constraints, iterative methods such as NMPC-based refinement \cite{DeDona2009}, genetic algorithms, particle swarm optimization \cite{6069411}, and optimal control formulations \cite{1657645} have been explored.

\subsubsection{Summary}
Trajectory generation methods, particularly for DF systems, can be broadly categorized into three types: trajectory generation without obstacles or state-input constraints, often solved using linear or quadratic programming, trajectory generation with obstacle avoidance but no state-input constraints, and trajectory generation with both obstacle avoidance and state-input constraints. Even in the simplest case, optimality is rarely guaranteed, and more complex scenarios introduce trade-offs between feasibility, computational efficiency, and adaptability.

\subsection{Contributions \& Paper Organization}

The contributions of this paper are summarized:
\begin{enumerate}  
\item \textbf{A new waypoint reward-based planning algorithm}: This approach enables flexible waypoint sequencing, allowing the robot to bypass lower-priority waypoints in favor of high-reward ones while ensuring all constraints are met. This is particularly valuable in real-world scenarios where constraints and limited resources make it impractical to complete all tasks, requiring intelligent prioritization to maximize mission success. The algorithm is demonstrated on multiple robotic platforms, including a ground vehicle, a quadruped, and a quadrotor. Additionally, benchmarking against Mixed-Integer Nonlinear Programming (MINLP) solvers shows that the proposed method scales more efficiently and outperforms alternative approaches.

\item \textbf{Efficient trajectory generation with obstacle avoidance and constraints}: We leverage clothoid curves and the differential flatness property to generate dynamically feasible trajectories while ensuring collision avoidance. Compared to traditional jerk and snap-based methods, this approach enhances smoothness and computational efficiency. Additionally, we introduce an effective constraint verification method that bypasses the high computational cost of nonlinear MPC, making it practical for combinatorial optimization problems that scale exponentially.

\item \textbf{Crossover mechanism for genetic algorithms in variable-length discrete spaces}: We propose a method to enable crossover between candidate solutions of different dimensions in discrete spaces using dynamic time warping, extended convex combination, and projection. Handling variable-length solutions in discrete optimization is inherently challenging, and this approach facilitates crossover in such scenarios. While demonstrated in waypoint-based planning, the method is general and can be applied to other discrete optimization algorithms.

\end{enumerate}

The rest of this paper is organized as follows: Section \ref{sec:problemF} presents the problem formulation; Section \ref{sec:algorithm} presents the proposed algorithm; Section \ref{sec:experiments} shows the numerical and experimental demonstrations; Section \ref{sec:performance} presents the algorithm benchmark and time complexity; Section \ref{sec:futurework} discusses the limitations and future extensions; Section \ref{sec:conclusions} concludes the paper and further summarizes the contributions.

\textbf{\emph{Notations.}}  We use bold symbols for vector variables, e.g. $\boldsymbol{x}, \boldsymbol{y} \in \mathbb{R}^n$. We do not bold sequences, but vector elements of sequences, e.g. if $s_1$ is a sequence of $n-$dimensional real vectors, the \( i \)-th element of the sequence is denoted as $\boldsymbol{s}_{1,i}\in \mathbb{R}^n$. $\boldsymbol{x} \leq \boldsymbol{y}$ indicates element-wise inequality. Similarly, for vector-valued functions $\boldsymbol{f}:\mathbb{R}^n \mapsto \mathbb{R}^m$, we use $\boldsymbol{y} = \boldsymbol{f}(\boldsymbol{x})$. 
Let $|\cdot|$ denote the Euclidean norm.
The union of sets \( A \bigcup B \) containing all elements that are in either \( A \) or \( B \), the set subtraction \( A \setminus B \) containing all elements that are in \( A \) but not in \( B \). The permutation Perm\((K)\) containing all permutations of the elements in  \( K \). The power set \( \mathcal{P}(A) \) is the set of all subsets of \( A \). Matrices are denoted by uppercase italic letters, such as $A$.

\section{Problem Formulation}
\label{sec:problemF}

\begin{figure}[!t]
\centering
\includegraphics[width=0.38\textwidth]{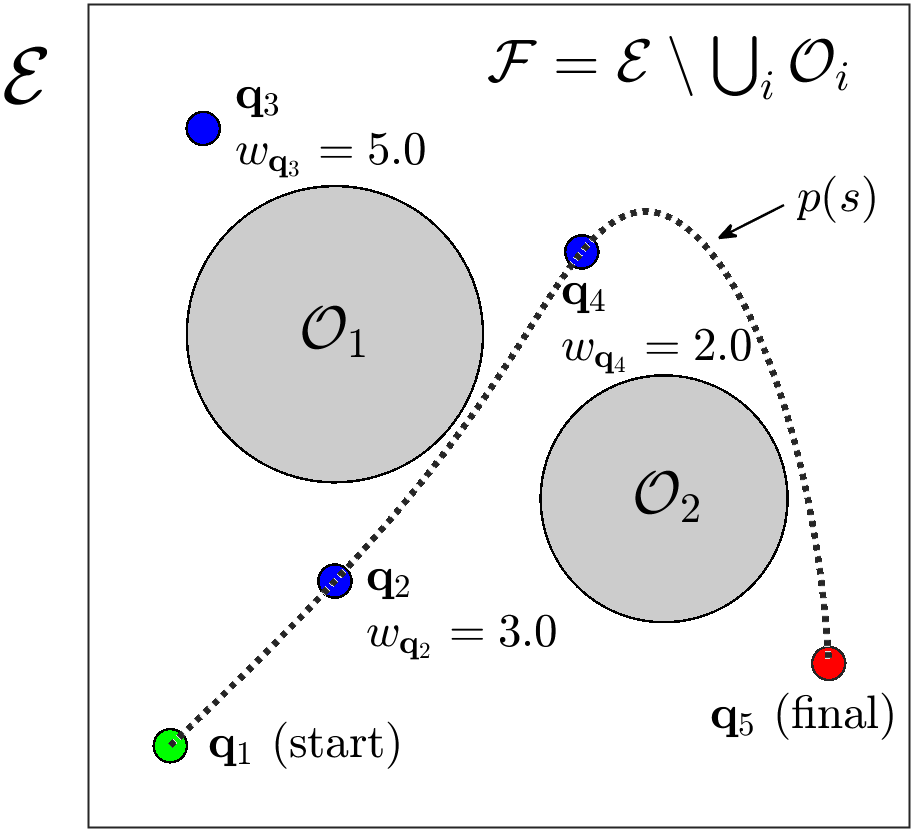}
\caption{Illustration of problem formulation elements: environment $\mathcal{E}$, obstacles $\mathcal{O}_i$, obstacle-free space $\mathcal{F}$, waypoints $\boldsymbol{q}_i$ with associated rewards $w_{\boldsymbol{q}_i}$, and obstacle free path $p(s)$ connecting the waypoints of the sequence $s=\{\boldsymbol{q}_1,\boldsymbol{q}_2,\boldsymbol{q}_4,\boldsymbol{q}_5\}$ in order.}
\label{fig:problem}
\end{figure}

Consider an environment $\mathcal{E}\subset\mathbb{R}^2$, where the elements of $\mathcal{E}$ represent positions. 
Within this environment, there are $N$ obstacles denoted as $\mathcal{O}_i \subset\mathcal{E}$, where the $i$-th obstacle $\mathcal{O}_i$ \hl{comprises} occupied points. Similarly, the obstacle-free space \( \mathcal{F} \) is defined as the set difference between the environment and the union of all obstacles, i.e.,
\begin{equation*}
\mathcal{F} = \mathcal{E} \setminus \bigcup_{i=1}^{N} \mathcal{O}_i .
\end{equation*}
There is a set of $\kappa$ human-given waypoints $\mathcal{Q}$ defined as
\begin{equation} 
\label{waypointset}
\mathcal{Q} = \{ \boldsymbol{q}_i \in \mathcal{F}: i=1,2,\dots \kappa \},
\end{equation}
which are assumed to be contained within the obstacle-free space $\mathcal{F}$, where $\boldsymbol{q}_1$ and $\boldsymbol{q}_\kappa$ represent the fixed initial and final points, respectively. Each waypoint $\boldsymbol{q}_i$, except for $\boldsymbol{q}_1$ and $\boldsymbol{q}_\kappa$, is assigned a reward $w_{\boldsymbol{q}_i} > 0$, representing the benefit of reaching that waypoint and influencing the optimal selection. Fig. \ref{fig:problem} illustrates this scenario. Similarly, the set of intermediate waypoints $\mathcal{Q}'$ is defined as the set difference between the waypoints set and the fixed initial and ending points $\mathcal{Q}' = \mathcal{Q} \setminus \{\boldsymbol{q}_1, \boldsymbol{q}_\kappa\}$.

The set of waypoint sequences $s\in\mathcal{S}$ that start from $\boldsymbol{q}_1$ and end at $\boldsymbol{q}_\kappa$, without repeating elements, is defined as follows:
\begin{equation*}
\mathcal{S} = \bigcup_{K \in \mathcal{P}(\mathcal{Q}')} \left\{ \left( \boldsymbol{q}_1,\sigma,\boldsymbol{q}_\kappa \right) \mid \sigma \in \text{Perm}(K) \right\},
\end{equation*}
where \( \mathcal{P}(\mathcal{Q}') \) represents the power set of \( \mathcal{Q}'\). For each subset \( K \) within \( \mathcal{P}(\mathcal{Q}')\), every possible permutation of \( K \) is considered, denoted as \( \text{Perm}(K) \). Following the example provided in Fig. \ref{fig:problem}, $K$ is any element of $\mathcal{P}(\mathcal{Q}')=\{\{\boldsymbol{q}_2\},\{\boldsymbol{q}_3\},\{\boldsymbol{q}_4\},\{\boldsymbol{q}_2,\boldsymbol{q}_3\},\{\boldsymbol{q}_2,\boldsymbol{q}_4\},\{\boldsymbol{q}_3,\boldsymbol{q}_4\},\{\varnothing\}\}$. For every permutation $\sigma\in\text{Perm}(K) $, it is defined a sequence that begins with \( \boldsymbol{q}_1 \), followed by the elements of \( \sigma \), and ends with \( \boldsymbol{q}_\kappa \). Here, $\left( \boldsymbol{q}_1,\sigma,\boldsymbol{q}_\kappa \right)$ denotes the concatenation of $\boldsymbol{q}_1$, $\sigma$, and $\boldsymbol{q}_\kappa$. The path $p(s)\subset\mathcal{F}$ denotes a set of obstacle-free points that forms a continuous curve and connects the elements of the sequence $s$ in order, as illustrated in Fig. \ref{fig:problem}.

Consider a robot with the following continuous dynamics and initial state:
\begin{equation}
\label{deqn_ex1}
\dot{\boldsymbol{x}}(t)=\boldsymbol{f}(\boldsymbol{x}(t),\boldsymbol{u}(t)) \text{   with   } \boldsymbol{x}(0)=\boldsymbol{x}_0,
\end{equation}
where $\boldsymbol{x}(t)\in\mathcal{X} \subset \mathbb{R}^n$ denotes the robot state, $\boldsymbol{u}(t) \in\mathcal{U} \subset\mathbb{R}^m$ denotes the control input, the vector function $\boldsymbol{f}:\mathcal{X} \times \mathcal{U}  \mapsto \mathbb{R}^n$ is assumed to be twice-differentiable, $t\in [0,t_f]$ denotes the time index with $t_f$ the finite-time horizon.
Assume there is a maximum mission time window $t_{\mathrm{max}}$, a maximum allowable travel distance $d_{\mathrm{max}}$, and a set of state and input constraints $\boldsymbol{\mathcal{G}}$ \eqref{saturations} that may include state and input derivatives, and control input bounds $\underline{u}\leq |\boldsymbol{u}|\leq \overline{u}$. The problem of interest is to find an optimal sequence of waypoints $s^*\in \mathcal{S}$ that solves the following optimization problem:
\begin{maxi!}|s|
{s \in \mathcal{S}}{g(s)=\sum_{\boldsymbol{q}_i\in s}w_{\boldsymbol{q}_i}}{\label{mainproblem}}{}
\addConstraint{\dot{\boldsymbol{x}}(t) = \boldsymbol{f}(\boldsymbol{x}, \boldsymbol{u}), \ \forall t \in [0, t_f]\label{constraintmap}}
\addConstraint{\boldsymbol{\mathcal{G}}(\boldsymbol{x},\dot{\boldsymbol{x}},\dots, \boldsymbol{u},\dot{\boldsymbol{u}},\dots) \leq \boldsymbol{0}, \ \forall t \in [0, t_f]\label{constraint2}}
\addConstraint{t_f \leq t_{\mathrm{max}}\label{constraint3}}
\addConstraint{d(p(s)) \leq d_{\mathrm{max}}\label{constraint4}}
\addConstraint{p(s) \subset \mathcal{F}\label{constraint5}, }
\end{maxi!} where $w_{\boldsymbol{q}_i}$ denotes the reward of reaching waypoint $\boldsymbol{q}_i$, $g(s)$ is the sum of the reached waypoints rewards, $d(p(s))$ is the length of the path $p$ that connects the elements of $s$ in order. Note that $s^*$ may be not unique. If multiple feasible optimal solutions are found, criteria such as minimum distance or time can be applied.

\section{Methodology}
\label{sec:algorithm}

In this section, we first discuss the key challenges that influence algorithm selection. Next, we detail the overall workflow of the proposed genetic algorithm, emphasizing the fitness function’s role in managing constraints \eqref{constraint2} - \eqref{constraint5} through the penalty method to guide candidate solutions toward optimality. We then describe the designed fitness function in Section \ref{sec:fitnessfunc}, which integrates the differential flatness property and incorporates constraint violation metrics employed in the penalty method. Following this, we outline the genetic operators that facilitate thorough exploration of the design space in Section \ref{sec:geneticoperators}. Additional details on trajectory generation are provided in Section \ref{trajectory}, and finally, we present the parameter selection study in Section \ref{sec:tuning}.

\begin{figure}[!t]
\centering
\includegraphics[width=0.26\textwidth]{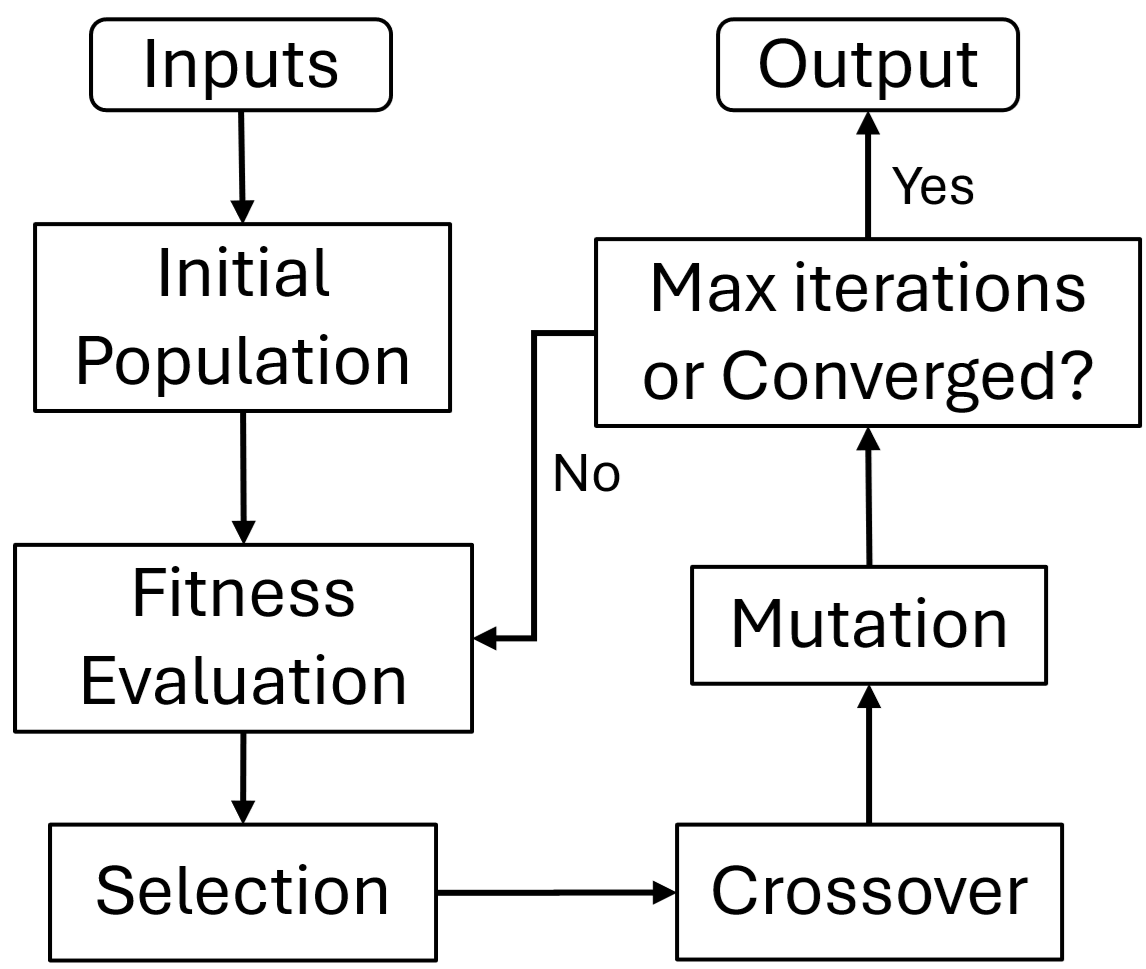}
\caption{Flowchart of the proposed algorithm based on classical Genetic Algorithms (GA).}
\label{fig:generalflowchart}
\end{figure}

As explained in the problem formulation, the decision variable $s$ is the sequence of the waypoints to be reached. Since there is no specific order to follow except for the initial and perhaps the final waypoint, the nature of the problem is combinatorial in the first level. As discussed in the introduction, the heuristic algorithms are usually employed in combinatorial optimization, and they have been theoretically and empirically proven to be a robust search method \cite{10.5555/534133}, despite having no theoretical guarantee of optimality in general.  Among the many heuristic optimization algorithms, genetic algorithms stand out for their versatility, making them particularly well-suited for handling variable-dimension discrete optimization problems. Moreover, genetic algorithm (GA) has demonstrated the best trade-off between accuracy and computational efficiency in solving closely related problems, such as the travelling salesman problem. The overall structure of the proposed algorithm follows the GA flowchart of Fig. \ref{fig:generalflowchart}.

The general workflow of the proposed genetic algorithm is outlined in Algorithm \ref{alg:alg1}, where convergence is achieved when the improvement in the best fitness value is below a small threshold over a set number of generations. The algorithm consists of multiple subroutines, which will be detailed in the following subsections. It takes the following inputs: the map, obstacle shape and position, the robot's dynamics, input saturation limits, distance and time constraints, and the maximum number of iterations. Additionally, any constraints on states, inputs, and their derivatives can be specified as needed.

Now we will outline the key parts of the proposed algorithm based on GA. First, the random initial population is generated and evaluated under the fitness function $h:\mathcal{S} \mapsto [1,\infty)$. This initial population is not required to lie in the feasible set. This fitness function measures the constraint fulfillment from a solution candidate along with the total collected reward, meaning it has a minimum when all the waypoints are reached, and all the constraints are fulfilled. In Section \ref{sec:fitnessfunc}, further details on the fitness function are discussed.

In the \texttt{Selection} process, we implement pre-crossover truncation removing the worst performing $T$ proportion of the population, followed by the elitism step, where the top individuals in the population are selected to survive directly into the next generation. Then, for each solution candidate $s_k$, a suitable partner is selected based on the fitness function values of each other candidate. Next, the \texttt{Crossover} function generates new individuals by combining the characteristics of both parents. Specifically, a portion of the offspring is produced using a warping-based approach (including projection and an extended convex combination) to align the sequences, while the remainder population is created via a random subsequence insertion method.

After the crossover, a \texttt{Mutation} function is applied to the new individual. This step is classical in all GA to increase design space exploration and avoid premature convergence to local minima. Then, the fitness evaluation is performed again, and the loop is repeated until it reaches either the maximum iterations or the population convergence.


\begin{algorithm}[ht]
\caption{General Workflow}\label{alg:alg1}
\DontPrintSemicolon
\KwIn{$\mathcal{E},\mathcal{O},\mathcal{Q},\boldsymbol{f}(\boldsymbol{x},\boldsymbol{u}),\underline{u},\overline{u},t_\mathrm{max},d_\mathrm{max}, iter_\mathrm{max}$,  $iter=0, h^* = 1e20$}
Random initial population $\{s_k : k=1,2,...,c\}$\;
\For{$k = 1$ \KwTo $c$}{
    Fitness function $h_k \gets h(s_k)$\;
    \lIf{$h_k < h^*$}{
        $s^* \gets s_k, \ h^* \gets h_k$
    }
}
\While{$\text{iter} < iter_\mathrm{max}$}{
    \For{$k = 1$ \KwTo $c$}{
        $\{\Bar{s}_k, s_k\} \gets \text{Selection}(s_k)$\;
        $s_{kb} \gets \text{Crossover}(\Bar{s}_k, s_k)$\;
        $s_k \gets \text{Mutation}(s_{kb})$\;
    }
    \For{$k = 1$ \KwTo $c$}{
        Fitness function $h_k \gets h(s_k)$\;        
        \lIf{$h_k < h^*$}{
        $s^* \gets s_k, \ h^* \gets h_k$
        }
    }
    $iter \gets iter + 1$\;
    \lIf{Converged}{
        $iter \gets iter_\mathrm{max}$ 
    }
}
\Return{$s^*$}
\end{algorithm}



\subsection{Fitness Function}
\label{sec:fitnessfunc}

In this section, we provide a detailed explanation of the proposed fitness function (Fig. \ref{fig:fitnessfunction}), as implemented in lines 3 and 11 of Algorithm \ref{alg:alg1}. The fitness function steers the evolution of the population through the design space by evaluating each candidate solution's performance relative to the optimal criterion, while also ensuring constraint satisfaction. To enforce constraint fulfillment, we employ the penalty method, which incorporates constraint violations into the evaluation process. The problem formulation \eqref{mainproblem} involves a single objective of maximizing the collected rewards with several constraints. Accordingly, a fitness function $h(s):\mathcal{S} \mapsto [1,\infty)$ is designed to have a minimum $h=1$ whenever $g(s)=\sum_{i=2}^{\kappa-1} w_{\boldsymbol{q}_i}$, i.e., $s$ reaches all the intermediate waypoints, and when all the constraints are also satisfied. Note that we expect a solution that minimizes the fitness function. Then $h(s)$ takes the form
\begin{equation}\label{fitnessfunction}
h(s)=\frac{g_\mathrm{max}}{g(s)}+R(s),
\end{equation}
where $g_\mathrm{max}=\sum_{i=2}^{\kappa-1} w_{\boldsymbol{q}_i}$ is the sum of all the available rewards, $g(s)$ is the sum of the reached waypoint rewards $w_{\boldsymbol{q}_i}$. $R(s):\mathcal{S} \mapsto [0,\infty)$ is the penalty function which is zero whenever the constraints are satisfied, defined as
\begin{equation}
\label{penalty}
R(s)=\sum_i^M\alpha_iV_i(s),
\end{equation}
where $M$ is the number of penalties, $\alpha_i$ is the weight for the \( i \)-th penalty, and $V_i$ is the violation/deviation measure for the \( i \)-th constraint. One of the main keys of the current proposal lies in how each violation measurement $V_i$ is performed.

The flowchart of the fitness evaluation is illustrated in Fig. \ref{fig:fitnessfunction}, where $s$ is the candidate sequence of waypoints, $A$ is the linear piecewise obstacle-free path connecting $s$ in order, $p_d$ the sampled smooth obstacle-free path connecting $s$ in order, $t_d$ the timestamp of each point in $p_d$, $x_d$ is the discrete sampled state, $u_d$ the discrete sampled input, $R(s)$ is the penalty function, and $h(s)$ is the fitness function value for the given sequence $s$.
This process is described as follows:

\begin{figure}[!t]
\centering
\includegraphics[width=0.30\textwidth]{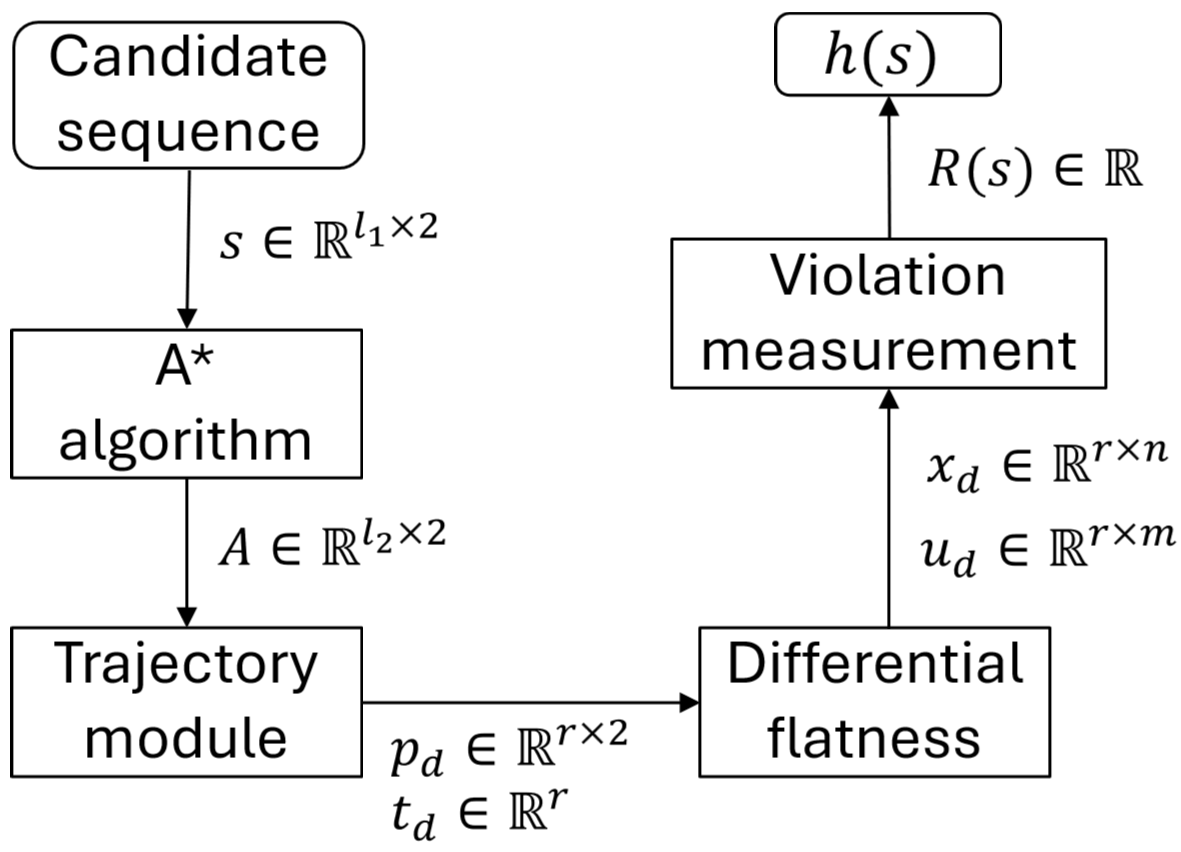}
\caption{Fitness function flowchart.}
\label{fig:fitnessfunction}
\end{figure}

\noindent \textbf{Step 1:} Given the sequence of waypoints in $s$, the path that connects these waypoints in the obstacle-free space is solved by the classic $A^*$ algorithm. It returns a set of extra grid coordinates between the intermediate waypoints that depend on the grid resolution of the map.\\
\textbf{Step 2:} A trajectory that goes through the $A^*$ solution is found by the trajectory module. It returns the time and the path $p(s)$. \\
\textbf{Step 3:} The trajectory is mapped into the state and input space using the DF property. It returns $\boldsymbol{x}(t), \boldsymbol{u}(t)$, and \hl{their} derivatives, from an open loop control scheme. \\
\textbf{Step 4:} The state, input, \hl{their} derivatives, time, and the length of $p(s)$ are compared against the constraint values. The violation measurements $V_i(s)$ are \hl{computed based on the magnitude and duration of the violations.}. Further details are explained in Section \ref{sec:violation}. \\
\textbf{Step 5:} The fitness function $h(s)$ is calculated based on $g(s)$ and $R(s)$, where suitable scale parameters $\alpha_i$ are selected.

\subsubsection{Differential Flatness}
\label{sec:DF}

As shown in Fig. \ref{fig:fitnessfunction}, the differential flatness (DF) is employed in the fitness function to map from a geometric trajectory to states and inputs. In this subsection, we introduce the concept of DF \cite{FLIESS1992159,doi:10.1137/S0363012995274027}. There is a particular type of nonlinear system that allows explicit algebraic trajectory planning, called differentially flat systems. It allows generating and tracking aggressive trajectories even for nonholonomic systems \cite{https://doi.org/10.48550/arxiv.2109.01365}, mainly because the trajectory planning can be done with methods designed for fully actuated systems. In other words, the path generation can be performed without particular concern about the dynamics. More efficient methods for optimal control can be developed for these systems \cite{BEAVER2024111404}.

Given the dynamics \eqref{deqn_ex1}, we call this system Differentially Flat (DF) if we can find outputs $\boldsymbol{y}\in\mathbb{R}^p$, called flat outputs, of the following form:
\begin{subequations}
\begin{align}
\boldsymbol{y}&=\boldsymbol{\xi}(\boldsymbol{x},\boldsymbol{u},\dot{\boldsymbol{u}},\dots, \boldsymbol{u}^{(\alpha_1)}),   \\
\boldsymbol{x}&=\boldsymbol{\gamma}_0(\boldsymbol{y},\dot{\boldsymbol{y}},\dots, \boldsymbol{y}^{(\alpha_2)}),   \\
\boldsymbol{u}&=\boldsymbol{\gamma}_1(\boldsymbol{y},\dot{\boldsymbol{y}},\dots, \boldsymbol{y}^{(\alpha_3)})  
\end{align}
\end{subequations}
with finite $\alpha_1,\alpha_2,\alpha_3$, and $(\boldsymbol{\gamma}_0,\boldsymbol{\gamma}_1)$ together are called the endogenous transformation. More precisely, there is a diffeomorphism between the flat output space and its derivatives, to the states and inputs. In the remainder of the article, we assume that system \eqref{deqn_ex1} is DF. However, this assumption is not as restrictive as it may seem, as DF systems include controllable linear systems, nonlinear systems that are feedback linearizable (by possible dynamic time-dependent feedback), quadrotors \cite{Zhou2014VectorFF, articlenice}, such as ground vehicles \cite{Dhaouadi2013DynamicMO}, robotic manipulators \cite{BEAVER2024111404}, and fixed-wing aircraft \cite{articlenice}. We also show how a quadruped robot can be seen as a flat system in a simplified framework in Section \ref{sec:quadruped}.

\subsubsection{Constraints Violation Measurement}
\label{sec:violation}

As explained in Section \ref{sec:fitnessfunc}, the constraints are addressed by the penalty method \eqref{penalty}, where each \( i \)-th constraint violation is measured by $V_i$ and weighted by $\alpha_i$. The idea is to increase (penalize) the value of the fitness function for those candidates who do not satisfy a constraint, and the penalty is more significant when the violation is greater or \hl{holds for a} longer time. 

Now we define the mission time window $t_\mathrm{max}$ as the first constraint, and the maximum distance $d_\mathrm{max}$ violation as the second one, then the corresponding violation measurements $V_1$, $V_2$ are calculated using the following two equations:

\begin{equation*}
V_1 = \begin{cases}
0 &\text{if $t_f\leq t_\mathrm{max}$}\\
\frac{t_f-t_\mathrm{max}}{t_\mathrm{max}} &\text{if $t_f> t_\mathrm{max}$}
\end{cases},
\end{equation*}

\begin{equation*}
V_2 = \begin{cases}
0 &\text{if $d(p(s))\leq d_\mathrm{max}$}\\
\frac{d(p(s))-d_\mathrm{max}}{d_\mathrm{max}} &\text{if $d(p(s))> d_\mathrm{max}$}
\end{cases},
\end{equation*}
where $d(p(s))$ is the length of the path $p(s)$ returned by the trajectory module for the candidate $s$, and $t_f$ is the final time given by the time parameterization described in Section \ref{trajectory}. Note that if the vehicle does not have a minimum velocity to fulfill, the mission time window is always satisfied based on the Section \ref{trajectory} method, and the time window violation may turn into a saturation input violation.

Next, the obstacle-free constraint should be guaranteed by the trajectory module. Depending on the dimension of the configuration space and the number of waypoints, any trajectory iteration within the fitness evaluation should be avoided to keep the computational cost reasonable. Consequently, we propose a measurement based on the length of the path \hl{that is} not in the obstacle-free space $\mathcal{F}$; denoting the obstacle-free constraint as the third constraint:
\begin{equation*}
V_3 = 1-\frac{\int_{0}^{t_f} B_{\{p(\tau)\subset \mathcal{F}\}} |\dot{p}(\tau)| \, d\tau}{d(p(\tau))},
\end{equation*}
where $p(\tau)$ is the path parameterized by the time $\tau$, $B_{\{p(\tau)\subset \mathcal{F}\}}$ is the binary function that equals $1$ if $p(\tau) \subset \mathcal{F}$ and $0$ otherwise, $|\dot{p}(\tau)|$ is the magnitude of the derivative of the trajectory respect to time, giving the local length, and $d(p(\tau))$ is the total length of the path. Note $V_3$ is zero if the entire path lies on the obstacle-free space, and one if no point of the path is within $\mathcal{F}$. Since this measurement varies between $0$ and $1$, a relatively high penalty weight $\alpha_3$ can be selected. 

The concept behind the violation measurement $V_3$ is to eliminate the need for iterative adjustments during trajectory generation. Specifically, if a small segment of the path is found to be colliding, the resulting penalty will be minor. If this path is selected as optimal, further refinement can be applied to adjust the trajectory.

Finally, for each inequality outlined in constraints \eqref{constraint2} - \eqref{constraint5}, a violation measurement 
$V$ is established. This approach is analogous to the obstacle-free measurement. If computational speed is a priority (resulting in no iterations within the trajectory module), or if no feasible solution is found even after trajectory iterations, more substantial penalties are imposed for larger segments of the trajectory that violate the constraints. As discussed in Section \ref{trajectory}, these constraints will be assessed over a finite number of time samples.

Notice that different types of constraints require different violation measurements. For an input saturation constraint of the shape $|\boldsymbol{u}(t)|\leq \overline{u}$, we propose the following violation measurement:
\begin{equation*}
\label{vu}
V_u = \frac{\int_{0}^{t_{f}} (|\boldsymbol{u}(\tau)|-\overline{u}) B_{\{ |\boldsymbol{u}(\tau)| > \overline{u} \}} d\tau}{t_{f}\overline{u}},
\end{equation*}
where $B_{\{ |\boldsymbol{u}(\tau)| > \overline{u} \}}$ is a binary function that equals $1$ when $|\boldsymbol{u}(\tau)| > \overline{u}$, and $0$ otherwise. This integral becomes a finite sum when the time sampling is performed. The maximum time and input saturation constraints are employed to regulate the values. In this way, the higher the time that an input constraint is not satisfied, the higher the penalty, and similarly, the greater the deviation from the constraint, the higher the penalty as well. Note that $\boldsymbol{u}(\tau)$ is found as a the function of the flat outputs. To handle different constraints, suitable functions have to be selected in the numerator of the violation measurement function, e.g., the difference is not proper for angle deviations.

\subsection{Genetic Operators}
\label{sec:geneticoperators}

The genetic operators—\texttt{Selection}, \texttt{Crossover}, and \texttt{Mutation}—implemented in lines 7, 8, and 9 of Algorithm \ref{alg:alg1}, are designed to balance exploration and exploitation of the search space, guiding the population toward the global optimum while ensuring constraint satisfaction through the fitness evaluation process. In genetic algorithm terminology, the goal is to drive the population's evolution toward this optimal solution. The \texttt{Selection} operator incorporates both pre-crossover truncation and elitism mechanisms to enhance the evolutionary process and preserve high-quality solutions.

\subsubsection{Selection}

To improve convergence and computational efficiency, we propose a pre-crossover truncation, wherein the worst-performing $T$ proportion of the population are removed before the crossover step. By eliminating these weaker individuals, the algorithm prevents poor genetic material from propagating into future generations while conserving computational resources for more promising candidates.

After the truncation step, we apply an elitism mechanism, a strategy in evolutionary algorithms that ensures the best-performing individuals are guaranteed to survive into the next generation. Specifically, we define an elite proportion $E$ of the population whose members are directly copied from one generation to the next without undergoing any modifications like crossover or mutation. By doing so, we preserve the most successful genetic material and prevent losing high-quality solutions due to random variation. This approach accelerates convergence and helps maintain consistent progress toward optimal or near-optimal solutions.

Finally, after accounting for individuals removed by pre-crossover truncation and those retained through elitism, the size of the offspring population is selected to maintain a constant total population. Parent selection for crossover is then conducted via the stochastic universal sampling method, favoring candidate solutions with higher fitness values.

\subsubsection{Crossover}
\label{sec:dtw}

We propose a two-strategy crossover method, where a subset of the offspring is generated using a warping-based approach, and the other subset comes from a random subsequence insertion procedure. We begin by describing the warping-based technique.

The warping-based crossover proceeds as follows. First, the selected parents $\Bar{s}_{k}$ and $s_k$ are aligned via a warping function. Next, an extended convex combination is performed using a random parameter. Finally, each element of the resulting crossed sequence is projected onto the set $\mathcal{Q}$, while also removing any duplicate entries.

To enable crossover between two parent sequences with different dimensions, we introduce a warping function. This function aligns the sequences $s_1=s_k$ and $s_2=\Bar{s}_{k}$, identifying an optimal alignment between them. The inputs and output of the warping process are defined as follows.

\begin{figure}[!t]
\centering
\includegraphics[width=0.35\textwidth]{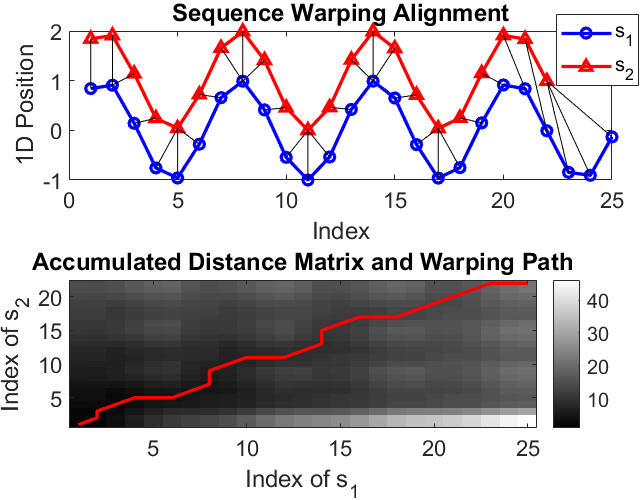}
\caption{Alignment example between the one-dimensional sequences $s_1$, $s_2$. The optimal alignment of the two sequences is \hl{found} by the warping path that follows the minimum distance matrix $D$ described by the \hl{color bar}.}
\label{fig:DTW}
\end{figure}

\begin{itemize}
    \item \textbf{Inputs}. $s_1$: a sequence of $l_1$ points in $\mathbb{R}^d$, $s_2$: a sequence of $l_2$ points in $\mathbb{R}^d$.  
    \item \textbf{Output}. $\texttt{Warped} \in \mathbb{R}^{a \times 2}$: a matrix of index pairs that defines the alignment between $s_1$ and $s_2$, with $a\leq l_1+l_2-1.$
\end{itemize}
Given that $\mathcal{E}\subset \mathbb{R}^2$, then $d=2$. The algorithm proceeds by constructing a distance matrix $D \in \mathbb{R}^{l_1 \times l_2}$, where the element $D_{i,j}$ represents the cumulative distance of the best path up to point $(i,j)$. The matrix is initialized as follows:
\begin{itemize}
    \item $D_{1,1} = |\boldsymbol{s}_{1,1} - \boldsymbol{s}_{2,1}|$,
    \item $D_{i,1} = D_{i-1,1} + |\boldsymbol{s}_{1,i} - \boldsymbol{s}_{2,1}|, \quad \text{for } i = 2, \ldots, l_1$,
    \item $D_{1,j} = D_{1,j-1} + |\boldsymbol{s}_{1,1} - \boldsymbol{s}_{2,j}|, \quad \text{for } j = 2, \ldots, l_2$,
\end{itemize}
where $\boldsymbol{s}_{k,i}$ is the \( i \)-th element of the sequence $k$. Subsequently, for each $i \in \{2, \ldots, l_1\}$ and $j \in \{2, \ldots, l_2\}$, the distance $D_{i,j}$ is computed as:
\begin{equation*}
    D_{i,j} = |\boldsymbol{s}_{1,i} - \boldsymbol{s}_{2,j}| + \min \{D_{i-1,j}, D_{i,j-1}, D_{i-1,j-1}\}.
\end{equation*}
The optimal path is then backtracked from $D_{l_1,l_2}$ to $D_{1,1}$, constructing the warping path \texttt{Warped} that minimizes the overall distance. A one-dimensional example is illustrated in Fig. \ref{fig:DTW}, where the optimal alignment of two sequences is found based on the distance matrix $D$ as described above. After the warping function $s_k$ and $\Bar{s}_{k}$ have the same length $a$, hence an extended convex combination is performed between each $d-$dimensional element of the warped sequences as following:
\begin{equation}
\label{convex}
\boldsymbol{s}_{k,i}=(1-\beta_i)\boldsymbol{s}_{k,i}+\beta_i\Bar{\boldsymbol{s}}_{k,i},\quad i=1,2,...,d,
\end{equation}
where $\beta_i\in[-0.15,1.15]$ is selected randomly for each crossover operation. Note that this definition allows a higher range and variability than a strict convex combination since it extends further than the convex hull.

\begin{figure}[!t]
\centering
\includegraphics[width=0.3\textwidth]{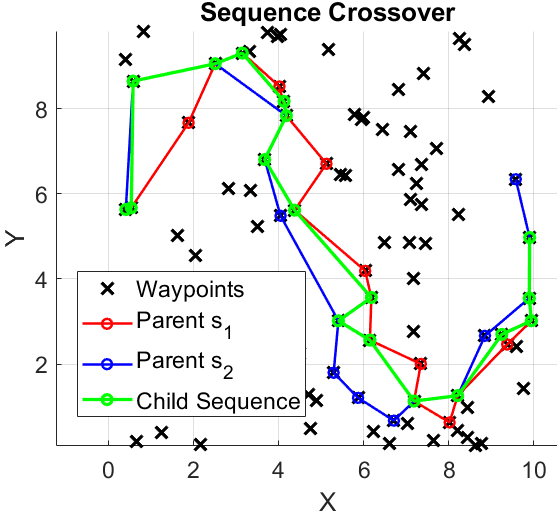}
\caption{Two-dimensional sequence crossover. Given two parent sequences of available waypoints, a new child sequence is obtained after the warping, combination, and projection.}
\label{fig:seqcross}
\end{figure}

Notice that, for each entry, the result of \eqref{convex} lies on the entire space $\mathbb{R}^d$. The optimization variable belongs to the finite set \eqref{waypointset}\hl{. Therefore, a } projection to the set $\mathcal{Q}$ is performed for each waypoint. This projection of each element of $s$ can be solved straightforwardly by finding the $\boldsymbol{q}_i$ that minimizes the Euclidean norm to that row. Finally, any duplicate waypoint in $s$ is removed, finding $s_{kb}$. This process is illustrated with an example in Fig. \ref{fig:seqcross}, where a child sequence $s_{kb}$ is obtained from the crossover of two parent sequences using the described warping, extended convex combination, and projection onto $\mathcal{Q}$.

The second crossover type is the random subsequence insertion. This method starts with a parent sequence $s_1$, from which a random subsequence is extracted. A second parent sequence $s_2$ is then selected, and elements in the random subsequence from $s_1$ are removed from $s_2$ using a set difference operation to ensure no duplicates. A random insertion point is chosen in this duplicate-free sequence, and the subsequence from $s_1$ is inserted at this location. The resulting sequence serves as a new candidate for the next generation.

Finally, the new generation is generated based on the proportion defined by the parameter $C$. When $C=0$, the entire new generation is derived from the subsequence crossover method, whereas $C=1$ indicates that all offspring are produced using the warping-based method. The selection of this parameter is discussed in Section \ref{sec:tuning}.

\subsubsection{Mutation}

The idea behind the \texttt{Mutation} operation is to introduce diversity into the sampled population and thereby avoid premature convergence. To this end, we propose a swap mutation that occurs with probability $p_m$ in which two waypoints within the candidate sequence are exchanged while the rest of the sequence, and then its length, remains unchanged. We exclude the first and, if specified, the last entry from the swap since they are fixed waypoints.

\subsection{Trajectory Generation}
\label{trajectory}

\subsubsection{Jerk/Snap Trajectory Challenges}
To plan motion through the desired waypoints, a method is needed that considers obstacle avoidance and state-input constraints, possibly including their derivatives. Since this is only one step in the fitness evaluation, it must run multiple times during the evolution process. This happens once for each individual in every generation. Therefore, we require a low-cost method, even if it is suboptimal.
The sub-optimality is not a severe limitation since most of the literature algorithms are sub-optimal, as discussed in Section \ref{sec:trajectory_lit}. Therefore, considering the computational cost, any method using iterative NMPC \cite{DeDona2009} should be avoided.

Although most methods in the literature use polynomial splines that minimize jerk or snap, this approach presents an additional challenge beyond what is discussed in Section \ref{sec:trajectory_lit}, which has not been widely recognized in the literature. Since the parameterization is based on time, a set of independent polynomial functions along each coordinate is created \hl{as} $x=P_x(t)$, $y=P_y(t)$, and such splines can not be represented in arc length form. This cannot be avoided since most trajectories will not fulfill the unique output requirement of the relation $(x,y)$ to be a function (for a coordinate $x$, there could be different $y$ values). Moreover, the curvature can not be controlled directly since the curvature is a relation between $x$ and $y$, but they are independent time polynomials $x=P_x(t)$, $y=P_y(t)$. The latter means we do not have control \hl{over} wavy shapes, including sharp knots, which may be undesirable for different reasons: dynamical feasibility, safety, collision avoidance, and efficiency. 

\begin{figure*}[tb!]
\centering
\subfloat[]{\includegraphics[width=0.25\textwidth]{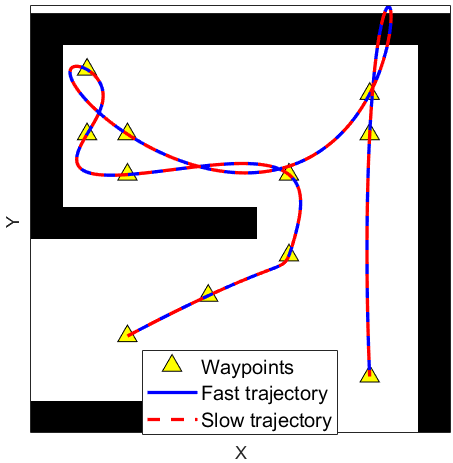}%
\label{fig:traj1a}}
\hfil
\subfloat[]{\includegraphics[width=0.33\textwidth]{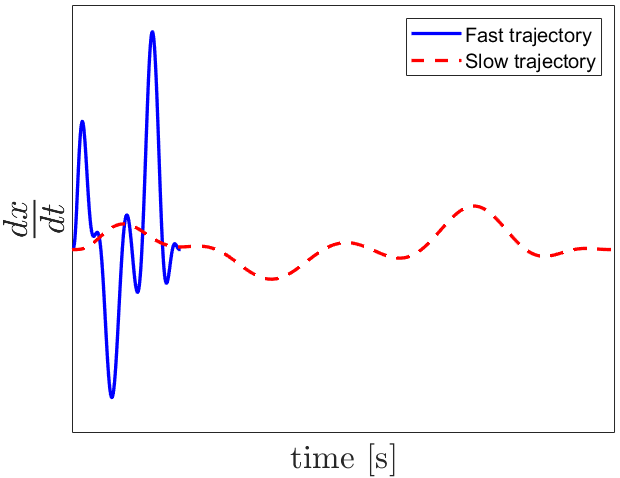}%
\label{fig:traj1b}}
\caption{Path of minimum snap with fast and slow \hl{timestamps}. Geometric space (a) (overlapped) and (b) time derivative of $x$. Slowing the trajectory does not alter the geometric path, thus failing to resolve obstacle collisions and sharp knots and turns.}
\label{fig:traj1}
\end{figure*}

We present an example of the aforementioned case. We test the classical snap minimization trajectory, with fixed \hl{timestamps} on ten waypoints, for two different time stamp sets. In the first case, the time difference between each waypoint is set as $1$ s; in the second case, it is $5$ s. As shown in Fig. \ref{fig:traj1a}, the two paths are exactly the same, even when the derivatives with respect to time are quite different, as shown in Fig. \ref{fig:traj1b}. Consequently, relaxing the time will not solve the issue.

\begin{figure*}[tb!]
\centering
\subfloat[]{\includegraphics[width=0.25\textwidth]{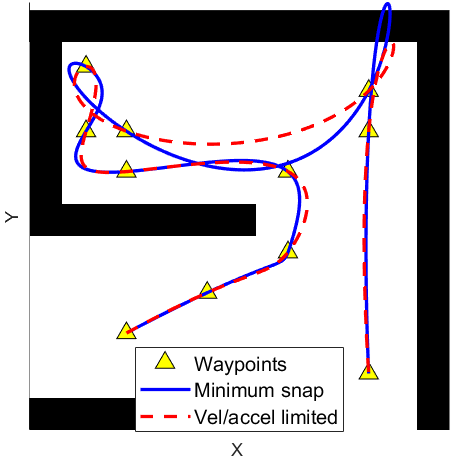}%
\label{fig:traj2a}}
\hfil
\subfloat[]{\includegraphics[width=0.33\textwidth]{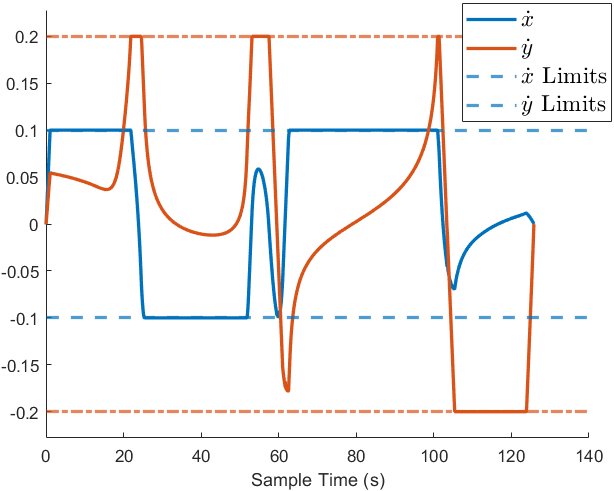}%
\label{fig:traj2b}}
\caption{Path of minimum snap and jerk with velocity and acceleration constraints. Geometric space (a) and (b) minimum jerk time derivative of $x$ and $y$ with the respective upper and lower bounds. Constraining velocities to minimize jerk and snap does not address sharp turns and knots, leading to collisions.}
\label{fig:traj2}
\end{figure*}

To further illustrate, we look for the time-optimal trajectory with velocities and accelerations constrained \cite{8338417} that minimizes the jerk. The result is shown in Fig. \ref{fig:traj2}. As expected, the velocities reach the bounds (Fig \ref{fig:traj2b}), and since it did not cross the boundary, it remains in the feasibility region. Despite the feasibility fulfilment with respect to the bounds, the described issues remain. Indeed, the path is not collision-free. Moreover, the shape differs from the minimum snap since it minimizes the jerk but this difference is not related to the rate bounds imposed to $|\Dot{x}|$ and $|\Dot{y}|$.

\subsubsection{Clothoid Curves vs Jerk/Snap Trajectories}

We propose a method to create a candidate trajectory for a given set of waypoints using piecewise clothoid curves, also called Euler spirals, instead of the classical polynomial basis that minimizes jerk or snap. The method defines a path segment between waypoints where curvature changes linearly with the distance, and the tangent at each $A^*$ waypoint is selected to minimize curvature discontinuities. This is a path without \hl{timestamps}. Since time allocation optimization can significantly increase computational cost \cite{articlenice}, we assume proportional \hl{timestamps} based on waypoint distance, time, and velocity. We use efficient clothoid interpolations from \cite{bertolazzi2015g1}.

\begin{figure*}[tb!]
\centering
\subfloat[]{\includegraphics[width=0.25\textwidth]{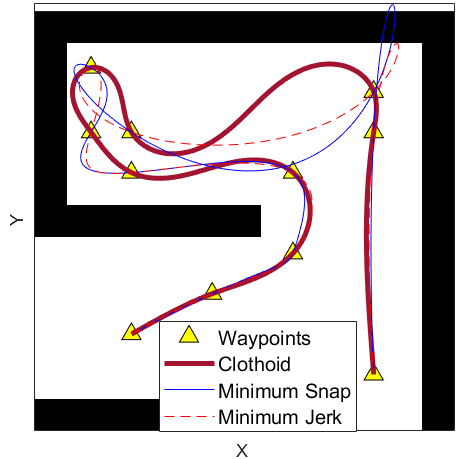}%
\label{fig:clothoida}}
\hfil
\subfloat[]{\includegraphics[width=0.25\textwidth]{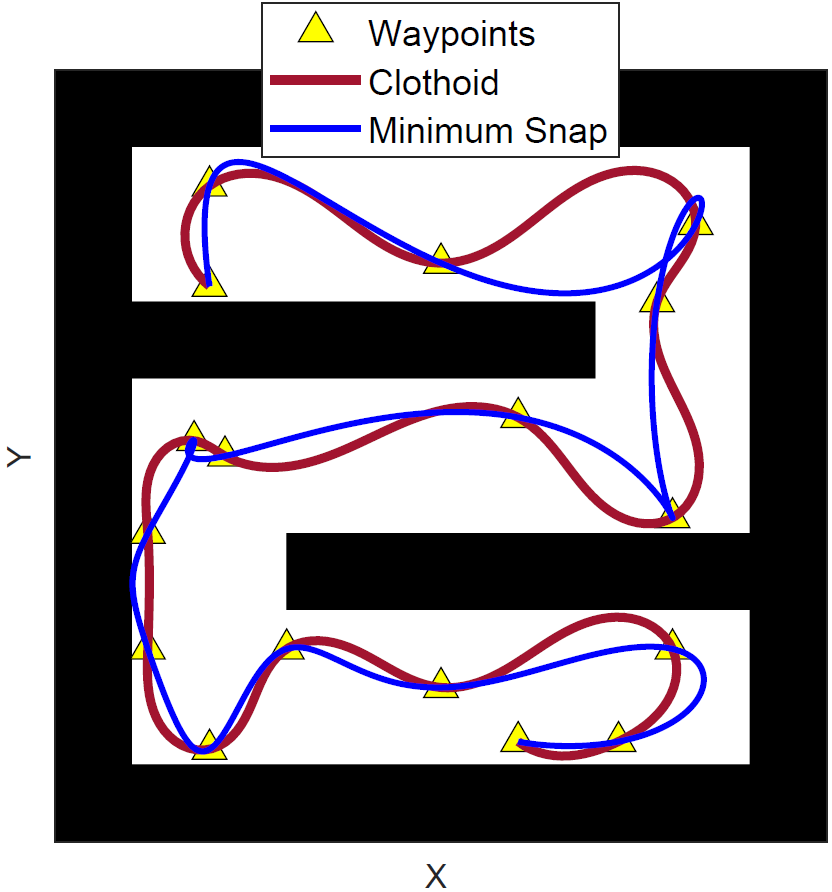}%
\label{fig:clothoidb}}
\caption{Clothoid trajectory comparison against minimum snap and jerk (a), and against minimum snap in a high-curvature scenario (b). The clothoid curve may require a larger turning radius\hl{,} but prevents sharp knots and undesired fast-wavy shapes that can be undesired for control systems.}
\label{fig:clothoid}
\end{figure*}

Figure \ref{fig:clothoida} presents a comparison between the clothoid trajectory and polynomial minimum jerk and snap trajectories, demonstrating a path that successfully avoids obstacles while mitigating sharp knots and high-curvature turns. Furthermore, Fig. \ref{fig:clothoidb} illustrates the behavior of the clothoid method in high-curvature scenarios. Although the clothoid approach effectively prevents the undesired wavy shapes that can adversely affect control systems, it may require a larger turning radius. This limitation, however, can be mitigated by introducing additional waypoints to enhance trajectory adaptability. If the jerk or snap needs to be limited in the clothoid parameterization, the same iterative method used in other trajectory generation approaches can be applied \cite{8338417}, specifically by slowing down the \hl{timestamps}.

The distance of a given path is calculated using this clothoid curve. Then, the time parameterization is performed depending on the mission time window constraint. Given time and length, an average constant speed is assumed to allocate the \hl{timestamps} into the path, finally finding the trajectory. This step is justified since the actuator limits are primarily determined by the distribution of time, and as the time goes to \hl{infinity}, the trajectory approaches a static state that is feasible for systems like quadrotors \cite{articlenice} and ground vehicles. Since we have a mission time window constraint, we use this value as the least aggressive possible trajectory. 

Suppose there is a minimum speed constraint, e.g., in fixed-wing aircraft, and the latter approach generates a speed below this limit. In that case, trajectory parameterization must be iteratively adjusted by reducing the final time until feasibility is reached. If no such constraint exists, no iterations are performed. The idea of generating an initial trajectory and adjusting it to satisfy constraints is also used in similar studies \cite{LAUMOND1994171, articlenice, 8338417}, often employing trajectory time reduction to ensure feasibility.

The collision avoidance of this algorithm lies in the $A^*$ solution and properties of clothoid curves. Since we are generating curves that change the curvature linearly, the reachable space or deviation distance is constrained by the length of the segment (bounded). After the clothoid curve generation, a collision check routine is employed. If there is a collision, an intermediate waypoint between two given points by $A^*$ is added (decreasing the length by two and then reduces the possible deviation), iteratively until the path lies entirely in the collision-free space, this process is guaranteed to reach a collision-free path since it converges to the piecewise linear solution from $A^*$, which is known to avoid collision.

\subsection{Parameter Tuning}
\label{sec:tuning}

We conducted 768 simulations to determine suitable values for various genetic algorithm parameters. The mutation probability was varied within the range $p_m = [0, 0.1, 0.2, 0.3]$, elitism within $E = [0.01, 0.05, 0.1, 0.2]$, and truncation within $T = [0, 0.1, 0.2, 0.3]$. Additionally, the crossover method parameter was tested in the range $C = [0, 1, 0.5]$, representing three configurations: offspring generated exclusively by the subsequence crossover method, exclusively by the warping-based method, or equally split between both methods, respectively. To ensure the parameter values perform well across scenarios of varying complexity, all combinations were tested in environments with $20$, $35$, $50$, and $65$ intermediate waypoints.

\begin{figure}
\centering
\includegraphics[width=0.38\textwidth]{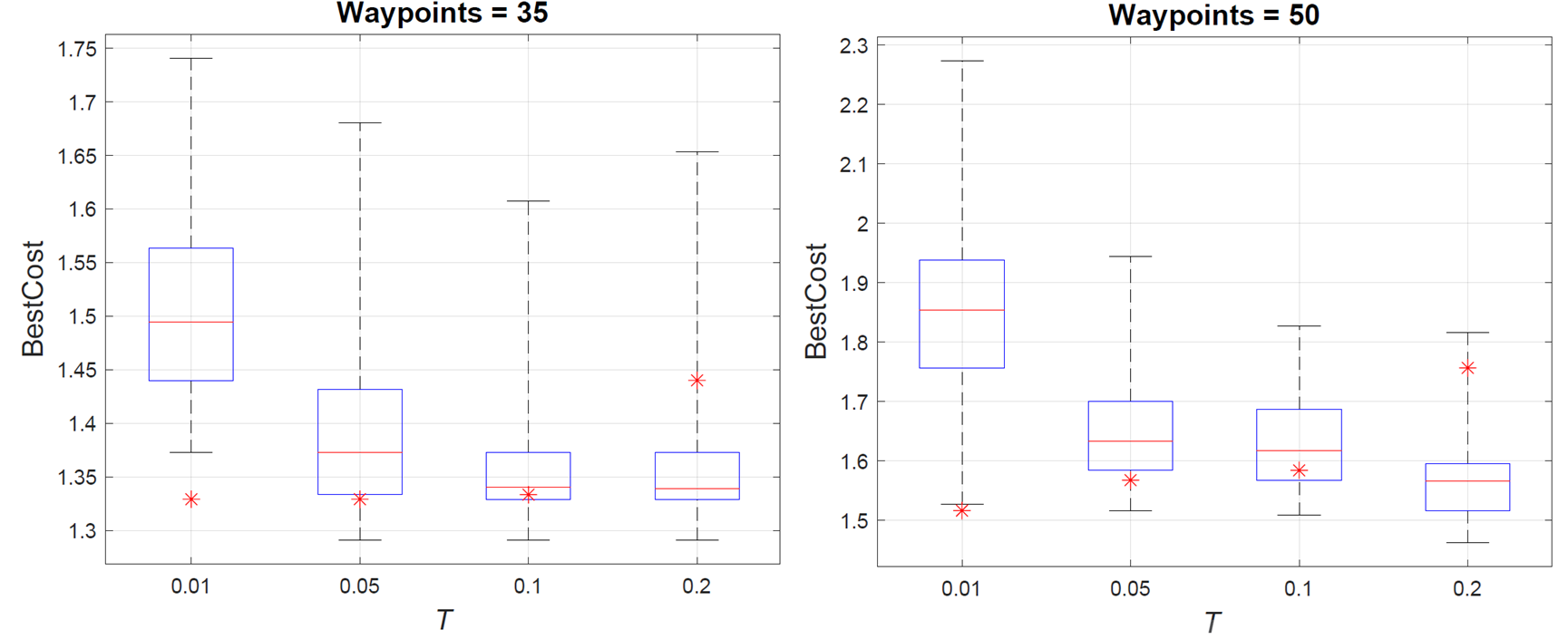}
\caption{Parameter tuning results for the truncation parameter $T$ in the $35$- and $50$-waypoint cases. In the $35$-waypoint scenarios, $T=0.1$ performed the best, while in the $50$-waypoint scenarios, $T=0.2$ achieved the best results.}
\label{fig:truncation}
\end{figure}

To determine the optimal set of parameters, we analyzed the results across all the scenarios for each number of waypoints. The trends for selecting individual parameters were not always consistent. For instance, as shown in Fig. \ref{fig:truncation}, the truncation parameter $T=0.2$ performed best in the $50$-waypoint scenarios but ranked as the second-best option in the $35$-waypoint scenarios. To address the parameter selection, we systematically identified and discarded the worst-performing parameter combinations for each waypoint scenario. This iterative process continued until only one parameter set remained, which demonstrated stable performance across all scenarios. While this final set did not always achieve the absolute best results in every case, it consistently avoided poor performance and excelled in several scenarios, making it a robust choice for varying complexities. The set of tuned parameters is $p_m=0.1$, $E=0.01$, $T=0.2$, $C=0.5$.

\hl{To evaluate the sensitivity of each genetic algorithm parameter with respect to the optimization performance, we employed a Random Forest regression model to estimate the relative importance of each parameter in predicting the final cost value.} 

\begin{figure}
\centering
\includegraphics[width=0.32\textwidth]{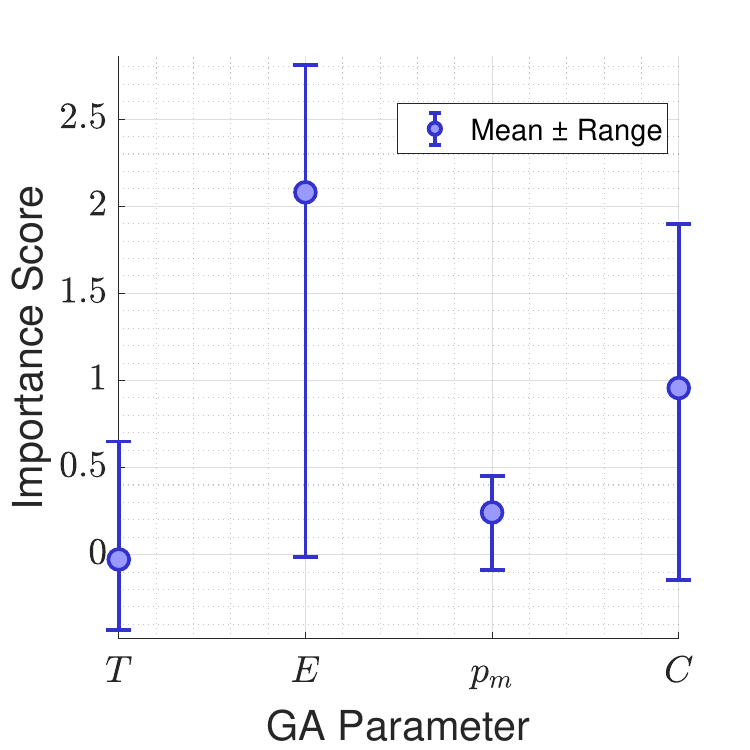}
\caption{\hl{Sensitivity of the best solution obtained by the algorithm to variations in genetic algorithm parameters: truncation $T$, elitism $E$, mutation rate $p_m$, and crossover method parameter $C$, evaluated across all waypoint configurations. The analysis indicates that elitism and the crossover method are the most influential parameters in determining performance, followed by the truncation parameter. The mutation rate exhibits the least overall impact.}}
\label{fig:sensitivity}
\end{figure}

\hl{For each number of waypoints 20, 35, 50 and 65, we trained a separate Random Forest model using the available GA parameter settings as predictors and the resulting \texttt{BestCost} as the response variable. The sensitivity of each parameter was quantified using the out-of-bag permuted predictor importance scores. This method computes the increase in prediction error when the values of a given predictor are randomly permuted, effectively breaking the association between that variable and the response. A larger increase in error indicates a more influential variable.

We repeated this process for each unique number of waypoints. For every parameter, we computed the mean, minimum, and maximum importance scores across problem sizes. These statistics are summarized in the error bar plot shown in Fig.}~\ref{fig:sensitivity}, \hl{where each marker represents the mean importance and the vertical bars indicate the range between the minimum and maximum values. The results suggest that elitism and the crossover method are the most influential parameters in determining the algorithm's performance, followed by the truncation parameter. In contrast, the mutation probability exhibits the lowest overall impact. For the population size $c$, we propose a linear growth with respect to the number of waypoints, following the rule} $c = \beta_p \kappa$ \hl{where} \(\beta_p \in [20, 40]\), \hl{based on similar randomized performance experiments.}

\section{Experiments}
\label{sec:experiments}

In this section, we present both simulation and hardware experiments. The simulations enable the handling of a greater number of waypoints and more complex obstacle configurations, making it ideal for demonstrating various scenarios where new waypoints can be introduced based on emerging zones of interest, increased rewards for specific waypoints, and changes in the mission time window. Additionally, the hardware experiments validate the feasibility of our approach in real control systems, highlighting the adaptability of the algorithm with more complex dynamics than the simulated ground vehicle, as the quadrotor and quadruped.

\subsection{Simulation Experiments}
\label{sec:simulations}
For the simulations, a differential-drive vehicle model is employed \cite{Dhaouadi2013DynamicMO}. It approximates a vehicle with a single fixed axle and wheels separated by a specified width, where the wheels can be driven independently. We assumed a 2D geometric space for simplicity and computational cost. The dynamics are described by
\begin{equation}\label{differentialdynamics}
\dot{\boldsymbol{x}} =
\begin{bmatrix}
\dot{x}_1 \\
\dot{x}_2 \\
\dot{x}_3
\end{bmatrix}
=
\begin{bmatrix}
\cos x_3 & 0 \\
\sin x_3 & 0 \\
0 & 1
\end{bmatrix}
\begin{bmatrix}
v \\
\omega
\end{bmatrix},
\end{equation}
where $x_1$ is the horizontal coordinate, $x_2$ is the vertical coordinate, $x_3$ is the vehicle heading respect to the horizontal axis, $v$ is the vehicle linear speed, $\omega$ is the vehicle heading angular speed. The vehicle speed and the angular speed are related to the left $\dot{\phi}_L$ and right $\dot{\phi}_R$ wheel angular speeds by the following equations:
\begin{subequations}
\begin{align}
v&= \frac{r}{2} (\dot{\phi}_R + \dot{\phi}_L), \\ 
\omega& = \frac{r}{d_v} (\dot{\phi}_R - \dot{\phi}_L),
\end{align}
\end{subequations}
where $r$ is the wheel radius and $d_v$ is the vehicle width. Since there is a one-to-one map between the wheel speeds and $v$ and $\omega$, we can consider either $\dot{\phi}_R$, $\dot{\phi}_L$ or the vehicle linear and angular speed as the control variables. To define the diffeomorphism required by the DF property, we consider the vector $[u_1 \ u_2]^T=[v \ \omega]^T$ as the control input, and $y_1$, $y_2$ as the flat outputs.

Then, the map from the outputs and \hl{their} derivatives to the state variables is defined by 
\begin{equation}
\begin{bmatrix}
x_1 \\
x_2 \\
x_3
\end{bmatrix}
=
\begin{bmatrix}
y_1 \\
y_2 \\
\tan^{-1}\left(\frac{\dot{y}_2}{\dot{y}_1}\right)
\end{bmatrix},
\end{equation}
and the map from the flat outputs and \hl{their} derivatives to the input is defined by 
\begin{equation}\label{diff2}
\begin{bmatrix}
u_1 \\
u_2
\end{bmatrix}
=
\begin{bmatrix}
\pm \sqrt{\dot{y}_1^2 + \dot{y}_2^2} \\
\frac{\dot{y}_1 \ddot{y}_2 - \ddot{y}_1 \dot{y}_2}{\dot{y}_1^2 + \dot{y}_2^2}
\end{bmatrix}.
\end{equation}
We also need the first derivative of $u_1$ defined by 
\begin{equation}
\Dot{u}_1 =\Dot{v} = \frac{\dot{y}_1 \ddot{y}_1 + \dot{y}_2 \ddot{y}_2}{\sqrt{\dot{y}_1^2 + \dot{y}_2^2}}.
\end{equation}
Note that we assume the vehicle is translating. Otherwise, $x_3$ and $u_2$ are not well defined by the diffeomorphism. Moreover, the angle $x_3$ has to be consistent with the sign of $u_1$, either forward or backward. Now we find the mapping from the states, input, and \hl{their} derivatives to the output and its derivatives, defined by
\begin{subequations}
\begin{align}
\begin{bmatrix}
y_1 \\
y_2
\end{bmatrix}
&=
\begin{bmatrix}
x_1 \\
x_2
\end{bmatrix},\\
\begin{bmatrix}
\dot{y}_1 \\
\dot{y}_2
\end{bmatrix}
&=
\begin{bmatrix}
u_1 \cos x_3 \\
u_1 \sin x_3
\end{bmatrix},\\
\begin{bmatrix}
\ddot{y}_1 \\
\ddot{y}_2
\end{bmatrix}
&=
\begin{bmatrix}
\Dot{u}_1 \cos x_3 - u_1u_2 \sin x_3 \\
\Dot{u}_1 \sin x_3 - u_1u_2 \cos x_3
\end{bmatrix}.
\end{align}
\end{subequations}

The geometrical environment is defined as $\mathcal{E}=[0,10]\times[0,10]\subset\mathbb{R}^2$. Three distinct obstacle sets are generated: maps $M_A$, $M_B$, and $M_C$. In $M_A$, 14 intermediate waypoints are created, each with a reward of $w=1$. The initial and final positions coincide, $\boldsymbol{q}_1=\boldsymbol{q}_{16}$, a common scenario in surveillance, mapping, and delivery. The input saturation is set to $\overline{u}_1=$ 1 m/s. Finally, a mission time window of $40$ s is established without a distance constraint.

\begin{figure*}[tb!]
\centering
\subfloat[]{\includegraphics[width=0.3\textwidth]{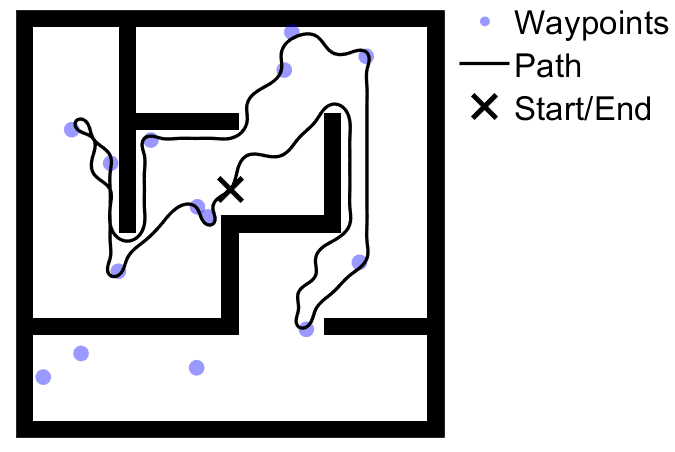}%
\label{fig_first_case}}
\hfil
\subfloat[]{\includegraphics[width=0.3\textwidth]{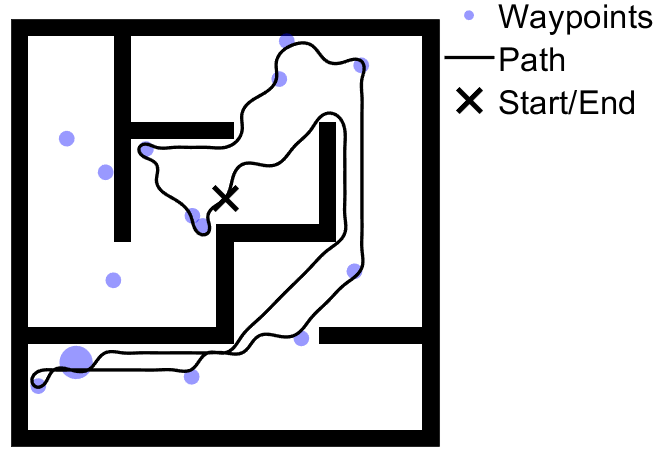}%
\label{fig_second_case}}
\caption{Map $M_A$ optimal path with $14$ intermediate waypoints and time constraint $t_\mathrm{max}=40$ s, with (a) $w=1$ for each waypoint, and a (b) deliberate increment on an waypoint located on the lower-left zone of the map.}
\label{fig_sim1}
\end{figure*}

Now we consider the same scenario, where one of the waypoints gets a deliberate increment in its weight up to $w=5$. Both results are shown in Fig. \ref{fig_first_case} and \ref{fig_second_case}. As a result of the increment of the lower left waypoint reward, the exploration in the upper left zone is reduced to invest that time in reaching the new highly rewarded waypoint. The actual $t_f$ of each case ($a$) and ($b$) is 36.8 s and 38.7 s, close to the mission time window, with rewards of 11 and 15, respectively, with no weights assigned to initial and final fixed points. 

\begin{figure*}[tb!]
\centering
\subfloat[]{\includegraphics[width=0.3\textwidth]{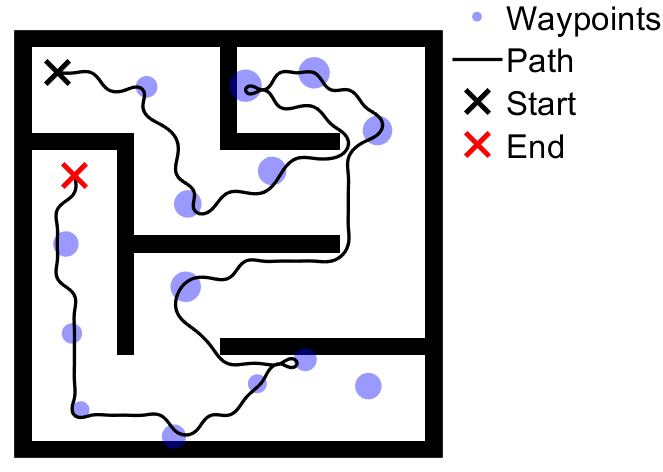}%
\label{fig_first_case2}}
\hfil
\subfloat[]{\includegraphics[width=0.3\textwidth]{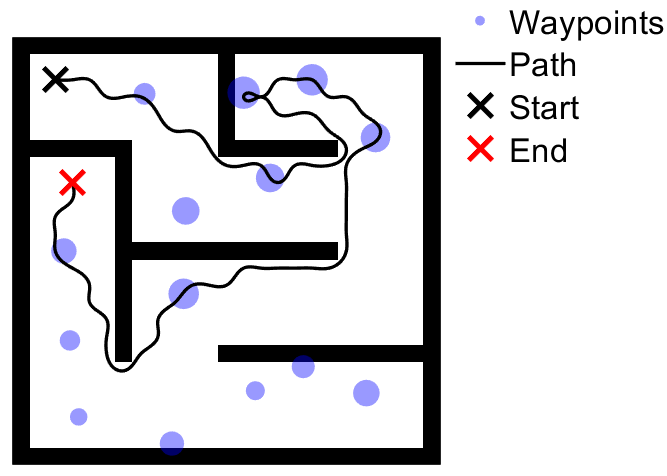}%
\label{fig_second_case2}}
\caption{Map $M_B$ optimal path with $14$ intermediate waypoints, with (a) time constraint $t_\mathrm{max}=40$ s, and (b) $t_\mathrm{max}=30$ s. Same waypoint reward values in (a) and (b).}
\label{fig_sim2}
\end{figure*}

For the second case of study, consider a map $M_B$, where the start and end positions differ, which is typical in scenarios where the robot's role is to connect a network of points, akin to a transit system connecting multiple distribution centers. The waypoint weights now have different values representing different importance levels of each location. The input constraints are the same as the scenarios on map $M_A$. Finally, a mission time window of $40$ s is set. Now, consider the same scenario, with a change in the time constraint to $30$ s, which can happen for mission requirements or just energy efficiency after noting the higher mission time window was unnecessarily large. The results are shown in Fig. \ref{fig_first_case2} and \ref{fig_second_case2}.

In map $M_B$ (Fig. \ref{fig_sim2}), the mission time window $t_\mathrm{max}$ decreases from $40$ s in case (a) to $30$ s in case (b). The actual $t_f$ are $39.78$ s and $29.84$ s, respectively. With the tighter constraint, it reaches fewer waypoints, prioritizing the most rewarded ones in the upper-right region of the map.

\begin{figure*}[tbp!]
\centering
\subfloat[]{\includegraphics[width=0.3\textwidth]{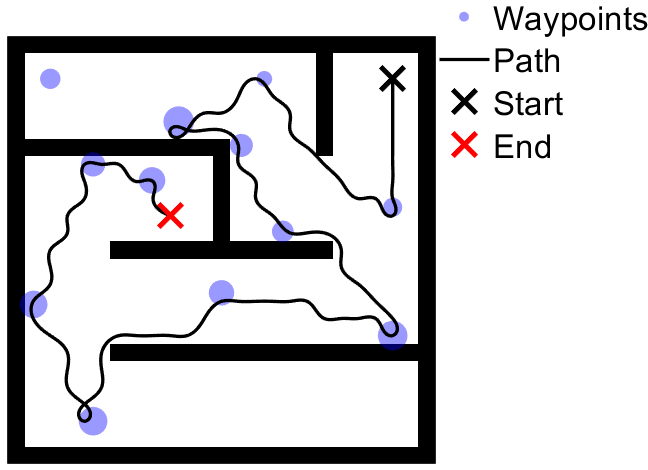}%
\label{fig_first_case3}}
\hfil
\subfloat[]{\includegraphics[width=0.3\textwidth]{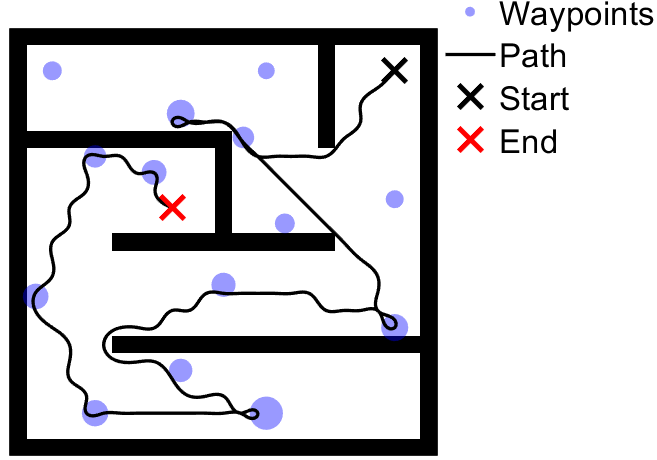}%
\label{fig_second_case3}}
\caption{Map $M_C$ optimal path with (a) $12$ and (b) $14$ intermediate waypoints. New highly rewarded waypoints were added (b) by the human to explore the lower zone. Same time window $t_\mathrm{max}=40$ s.}
\label{fig_sim3}
\end{figure*}

As the third study case, consider a map $M_C$ with different initial and final positions; 12 intermediate waypoints are distributed throughout the map. The mission time window is set as $t_\mathrm{max}=40$ s. Now, consider that there is a new interest in an unexplored zone that drives humans to add new highly weighted waypoints in the lower zone of the map. The solution evolution is shown in Fig. \ref{fig_sim3}. In this scenario, the new waypoints in the lower-right zone guide the trajectory toward the required exploration area, bypassing some waypoints in the upper and middle zones as a trade-off.

\subsection{Hardware Experiments}

To demonstrate the versatility of the proposed algorithm, we apply it to three dynamical systems: a ground vehicle, a quadrotor, and a quadruped. The ground vehicle and quadrotor are known DF systems, and we show how the quadruped can also be treated as DF using a kinematic control approach. The experiments take place in a $3$ m × $5$ m indoor space with two red rectangular obstacles. A motion capture system is used for localization and state estimation.  A video of the hardware experiments is available at \url{https://youtu.be/iQ9f3bVbYis} or in the supplementary materials. Each robot follows the trajectory generated by the proposed algorithm using its own low-level tracking controller. Technical details of the controllers are omitted as they are beyond the scope of this paper.

The ground vehicle is a classic example of a system where energy limitations primarily translate into distance constraints rather than time constraints, as most of the energy is consumed during motion, with communication and minimal standby power being relatively minor. In contrast, for a quadrotor, energy is directly tied to flight time, making the available mission time window the primary limiting factor in real-world scenarios. Finally, the quadruped demonstration serves to illustrate how this algorithm is still suitable for non-differentially flat systems in some scenarios.

\subsubsection{Ground Vehicle}


The first experiment uses the Jackal ground vehicle. It has been shown that the two-axis vehicle, considering the wheel's steering angle, is differentially flat \cite{sekhavat2001motion}. We assume the same DF ground vehicle model from Section \ref{sec:simulations} for simplicity. To perform planning in the configuration space, the geometric map is inflated by the vehicle’s maximum radius plus an additional safety margin, ensuring that planned paths maintain a safe buffer around the vehicle to prevent collisions with obstacles.

\begin{figure*}[tb!]
\centering
\subfloat[]{\includegraphics[width=2.8in]{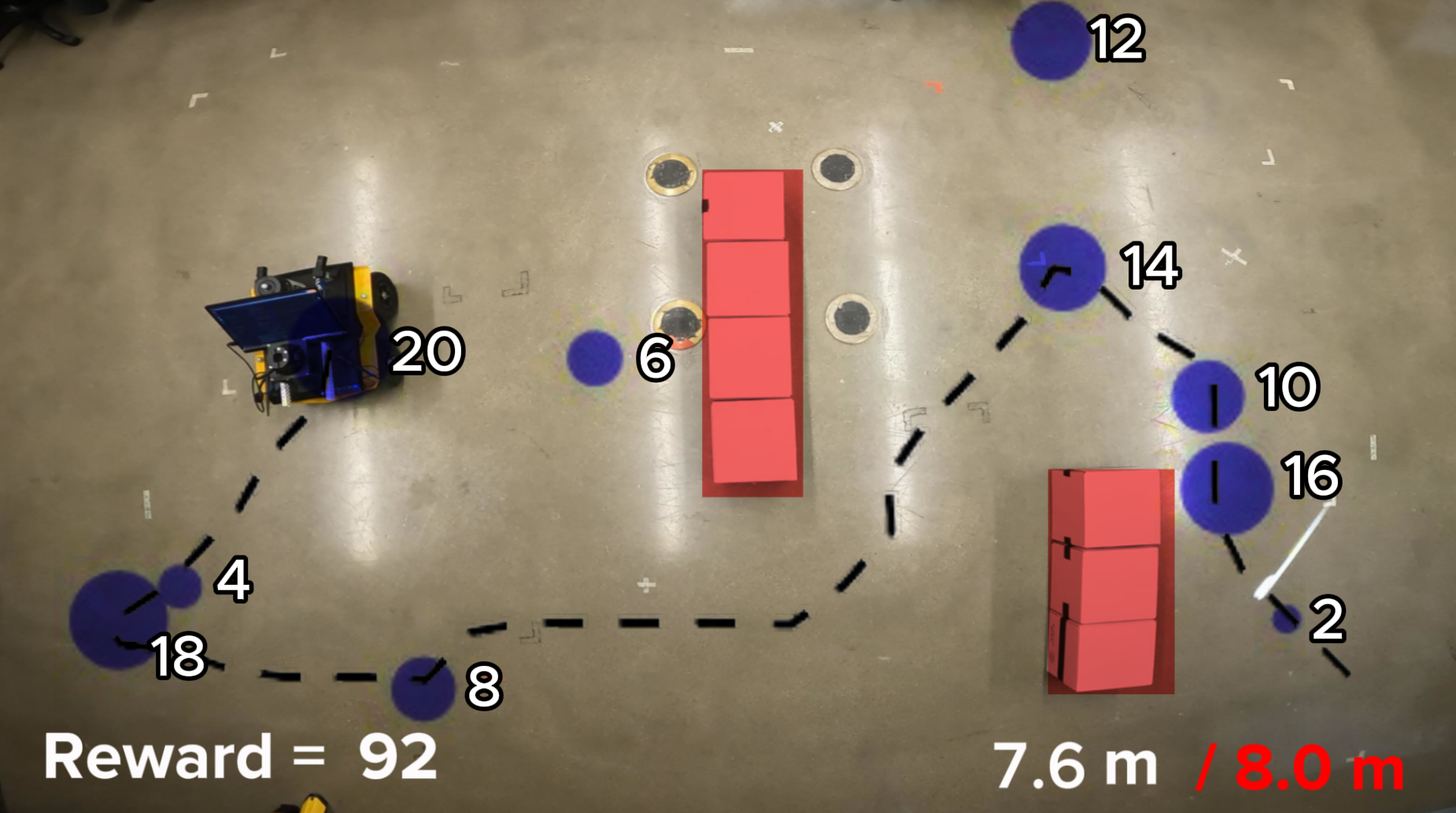}%
\label{fig:gvtraj}}
\hfil
\subfloat[]{\includegraphics[width=2.6in]{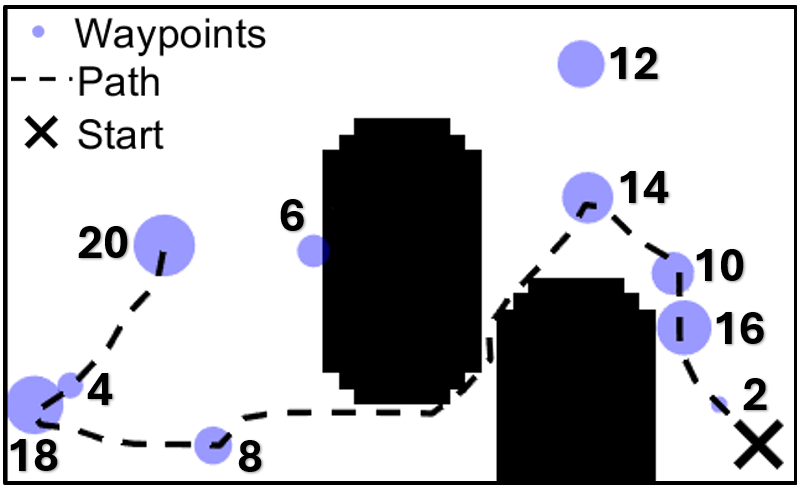}%
\label{fig:gvconfigspace}}
\caption{Optimal path in the ground vehicle experiment with a distance constraint and waypoint reward values. (a) geometric space path and (b) $x-y$ configuration space path. }
\label{fig:gvexperiment}
\end{figure*}

In this experiment, only the start position is fixed, and the final location is set free; ten intermediate waypoints are distributed through the map with rewards starting at $w_{\boldsymbol{q}_2}=2$ and increasing by increments of 2, reaching $w_{\boldsymbol{q}_{11}}=20$.
No mission time window is specified as a constraint. A maximum distance is selected as $d_\mathrm{max}=8$ m, along with input constraints $v_\mathrm{max}=$ 0.2 m/s, $\omega_\mathrm{max}=$ 0.5 rad/s. The experiment results and $x-y$ configuration space trajectory are shown in Fig. \ref{fig:gvexperiment}.

The maximum available reward is given by \( \sum_{i=2}^{11} w_{\boldsymbol{q}_i} = 110 \). Due to the distance constraint, only a subset of the waypoints can be reached, resulting in a maximum collected reward of \( 92 \). The result is illustrated by the trajectory shown in Fig. \ref{fig:gvconfigspace}. Hypothetically, instead of squeezing itself through two obstacles to reach the waypoint with reward 8, it reaches the waypoints of reward 14, 12, and 18, keeping the other waypoints in the same order. In this case, the total collected reward would increase to 96\hl{,} but the distance traveled would be 8.5 m $> d_{\mathrm{max}}$.

\subsubsection{Quadrotor}

Ref. \cite{Zhou2014VectorFF} proves that the quadrotor dynamics is differentially flat with flat output $\boldsymbol{\xi}=(\boldsymbol{x},\psi)^T\in \mathbb{R}^4$. The mapping from inputs and states to outputs can be found in Eqs. (7) - (10) of \cite{Zhou2014VectorFF}. Notably, as explained before, using the DF map eliminates the need for numerical integration of \eqref{constraintmap} to verify constraint fulfillment. Additionally, any constraint in \eqref{constraint2} - \eqref{constraint5} is transformed into a constraint on the flat outputs, also avoiding the numerical integration.

\begin{figure*}[tb!]
\centering
\subfloat[]{\includegraphics[width=3.0in]{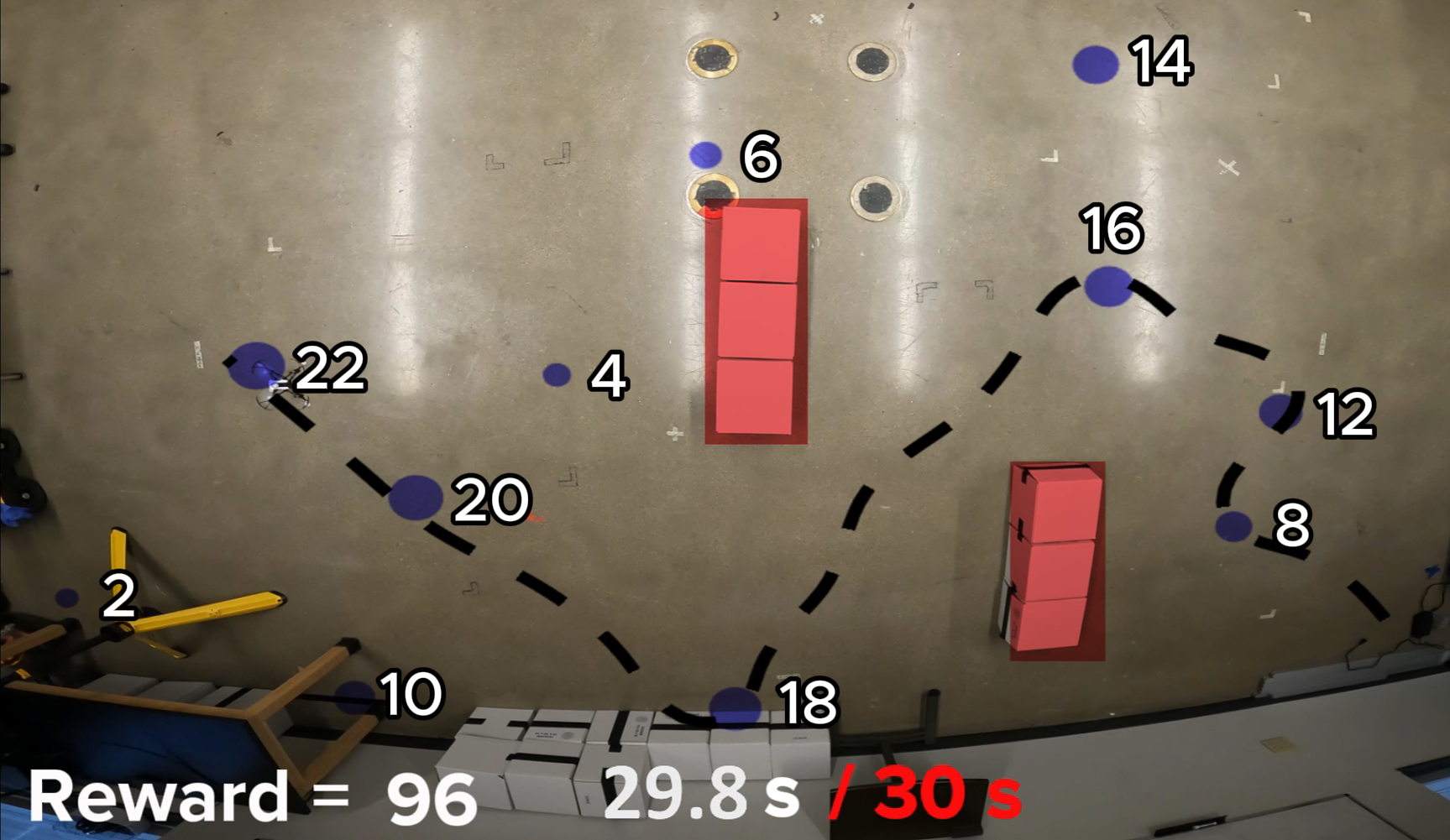}%
\label{fig:quadtraj}}
\hfil
\subfloat[]{\includegraphics[width=2.6in]{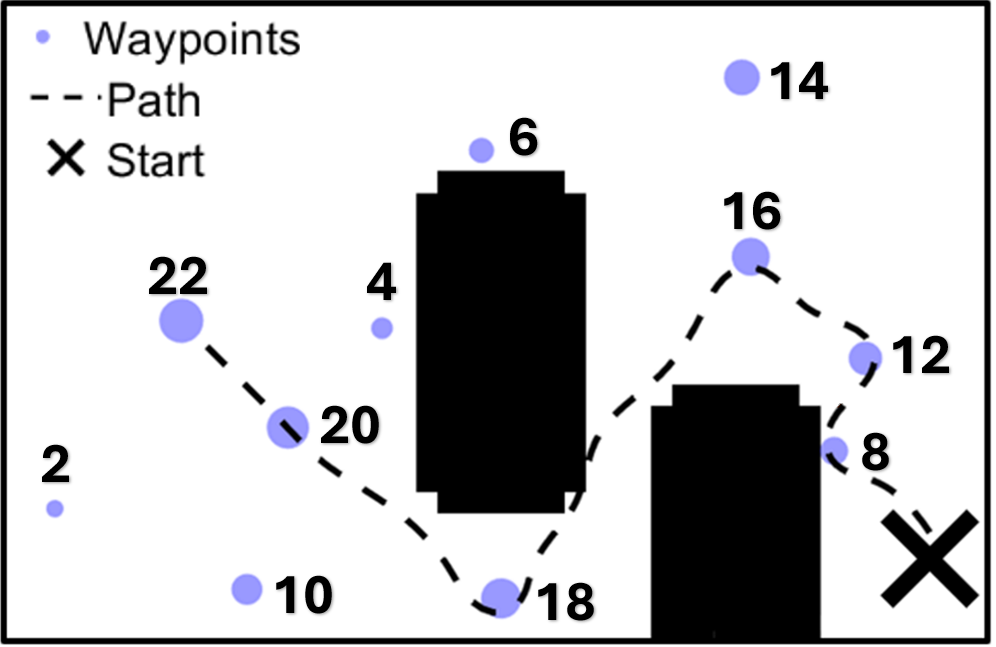}}%
\label{fig:quadconfig}
\caption{Optimal path in the quadrotor experiment with a time constraint and waypoint reward values. (a) geometric space path and (b) $x-y$ configuration space path.}
\label{fig:quadexp}
\end{figure*}

The geometric environment and obstacles are shown in Fig. \ref{fig:quadexp}. There is no distance constraint but a mission time window $t_\mathrm{max}=30$ s. The speed constraint is $|\dot{\boldsymbol{x}}| \leq 0.3$ m/s, and the maximum linear acceleration is $0.5$ m/s$^2$. In total, 11 intermediate waypoints are distributed, with rewards starting at \( w_{\boldsymbol{q}_2} = 2 \) and increasing in increments of 2, culminating in \( w_{\boldsymbol{q}_{12}} = 22 \). Only the initial position is specified, \hl{while the final position is left unconstrained}. The optimal trajectory, shown in Fig. \ref{fig:quadexp}, achieves a total reward of $96$ within a time of $29.8$ s.

\subsubsection{Quadruped}
\label{sec:quadruped}

\begin{figure}[!t]
\centering
\includegraphics[width=0.15\textwidth]{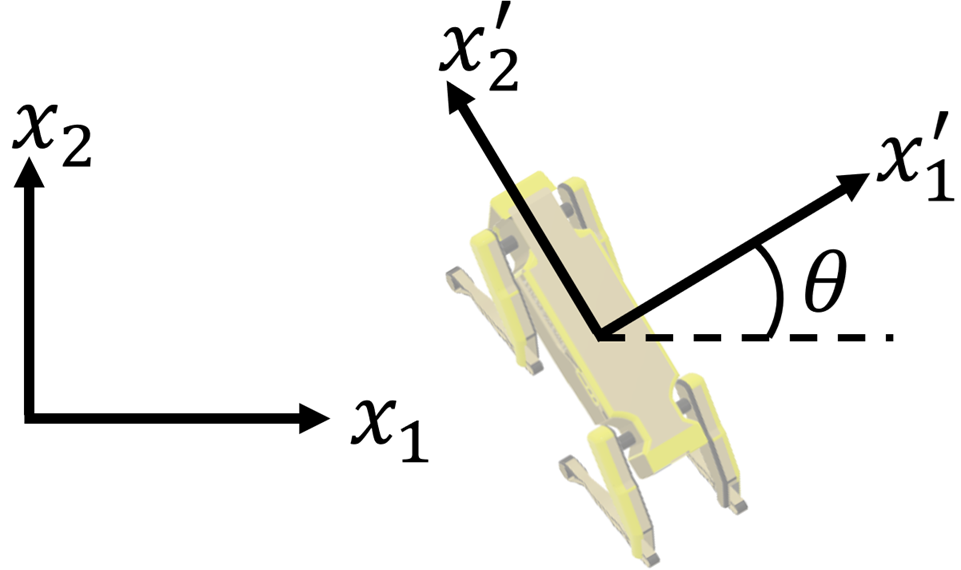}
\caption{Quadruped local and global frame.}
\label{fig:dogrobot}
\end{figure}

We demonstrate our approach using the Unitree Go1 quadruped robot. The robot is equipped with two front and two rear legs, each having three joints with specific angular constraints.
To the best of the authors' knowledge, no differentially flat mapping from the states and inputs of the robot's contact-aware multi-rigid-body dynamic model to its outputs has been presented in the literature.
Since the proposed algorithm emphasizes high-level task allocation and planning, we utilize a simplified kinematic model that takes forward-backward speed, left-right speed, and orientation as inputs, following a similar concept in the literature \cite{shamsah2023integrated}.
Although this simplification limits more complex planning, such as navigating uneven terrains \cite{fan2020geometric}, the use of hierarchical planning with models of varying fidelity remains effective for task planning \cite{shamsah2023integrated}.
In this high-level framework, we consider the state $\boldsymbol{x}=[x_1 \ x_2 \ \theta]^T$ in a global reference frame, as indicated in Fig. \ref{fig:dogrobot}, and we define the flat output as $\boldsymbol{y}=\boldsymbol{x}$. Then, the map from the flat output to the state is the identity, and the map from \hl{the flat} output and its derivatives to the flat inputs is
\begin{equation}  \begin{aligned}\label{eq:quadruped1}
\begin{bmatrix}
\dot{x}'_1 \\
\dot{x}'_2 \\
\omega
\end{bmatrix}
= \begin{bmatrix}
\cos\theta & -\sin\theta & 0\\
\sin\theta& \cos\theta& 0 \\
0& 0& 1
\end{bmatrix}
\begin{bmatrix}
\dot{x}_1 \\
\dot{x}_2 \\
\dot{\theta}
\end{bmatrix}+  \\
\dot{\theta} \begin{bmatrix}
-\sin\theta & -\cos\theta & 0\\
\cos\theta& -\sin\theta& 0 \\
0& 0& 0
\end{bmatrix}
\begin{bmatrix}
x_1 \\
x_2 \\
\theta
\end{bmatrix},
\end{aligned}
\end{equation}
where $[\dot{x}'_1 \ \dot{x}'_2 \ \omega]^T=[u_1 \ u_2 \ u_3]^T$ are the left-right, forward-backward, and rotational speeds in Fig. \ref{fig:dogrobot}. This completes the simplified quadruped flatness map used in the algorithm. 

\begin{figure*}[tb!]
\centering
\subfloat[]{\includegraphics[width=2.8in]{Fig1.png}%
\label{fig:dogtraj}}
\hfil
\subfloat[]{\includegraphics[width=2.4in]{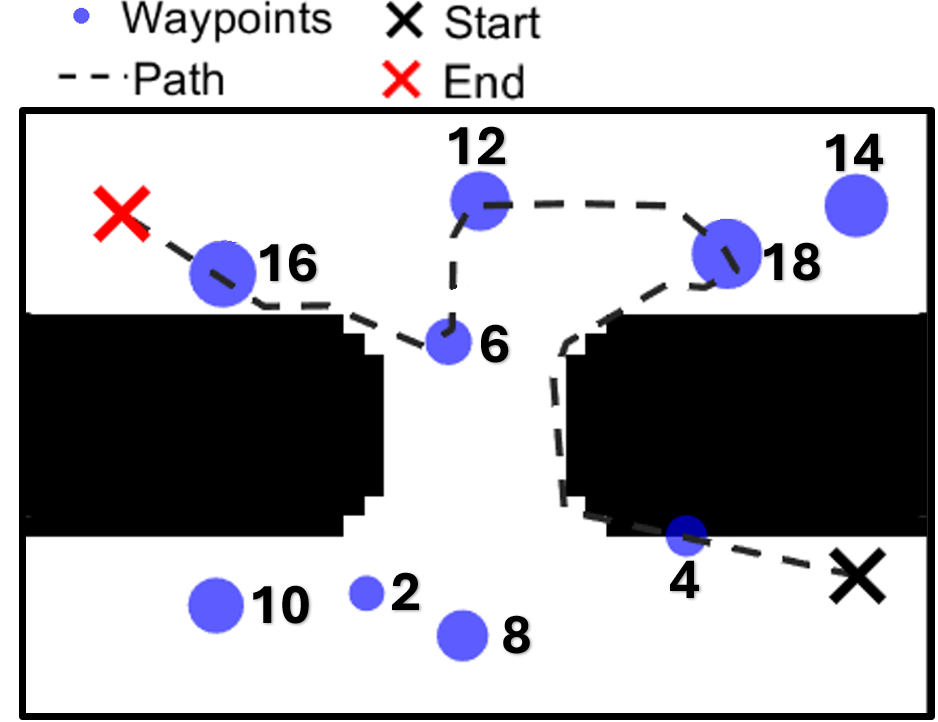}%
\label{fig:dogconfigspace}}
\caption{Optimal path in the quadruped experiment with a distance constraint and waypoint reward values. (a) geometric space path and (b) $x-y$ configuration space path.}
\label{fig:dogexp}
\end{figure*}

In total, nine intermediate waypoints are distributed across a new map shown in Fig. \ref{fig:dogconfigspace}, with rewards starting at $w_{\boldsymbol{q}_2}=2$ and increasing by increments of 2, reaching $w_{\boldsymbol{q}_{10}}=18$. An initial and final configuration are fixed, along with a maximum distance constraint of $7.5$ m. As a result of this constraint, the quadruped collected a total reward of $56$, as illustrated in Fig. \ref{fig:dogexp}, maximizing the collected reward while navigating collision-free.

\section{Algorithm Performance}
\label{sec:performance}

In this section, we evaluate the performance and scalability of the proposed algorithm through a benchmark study against available solvers and an analysis of its time complexity. Given that existing solvers are not well-suited to address the scale and complexity of the problem \eqref{mainproblem} directly, we instead optimize a more tractable fitness function \eqref{fitnessfunction}. To tackle this challenge, we develop two alternative formulations of the problem as Mixed-Integer Nonlinear Programming (MINLP) tasks and integrate them with two commercial solvers capable of handling MINLP: MATLAB's surrogate optimization and genetic algorithm (R2024b). Several factors make the use of existing solvers still challenging for this problem, including the computational difficulty of enforcing the no-duplicate constraint, the variable-dimensional nature of the optimization space due to the sequence's varying length, and the nonlinearities present in both the objective function and the constraints. We first present the benchmarking, followed by the complexity analysis.

\subsection{Benchmarking}

\subsubsection{Permutation-Based Integer Formulation}

In this approach, we define integer variables 
$z_{ij}\in\mathbb{Z}^{(\kappa-2)\times (\kappa-2)}$ as:
\[
z_{ij} =
\begin{cases}
1, & \text{if waypoint } i \text{ is assigned to position } j \\
0, & \text{otherwise}
\end{cases}
\]
for $i = 2, 3, \dots, \kappa -1$ (waypoints), and $j = 2, 3, \dots, \kappa -1$ (positions in the sequence). We exclude the fixed initial and final waypoints. \hl{To} have each waypoint assigned to at most one position, one can use $\sum_{j=2}^{\kappa -1} z_{ij} \leq 1, \forall i \in \{2, 3, \dots, \kappa -1\}$, and similarly for each position to have at most one waypoint assigned $\sum_{i=2}^{\kappa -1} z_{ij} \leq 1, \forall j \in \{2, 3, \dots, \kappa -1\}$. \hl{With these two constraints, we assure there are no repetitions, and the robot will reach a single waypoint at a time.} Moreover, since we want to force variable length sequences, we include the total number of assignments (intermediate waypoints) as a decision variable $\lambda$, and we constraint the waypoint assignments by $\sum_{i=2}^{\kappa -1} \sum_{j=2}^{\kappa -1} z_{ij} - \lambda \leq 0$. Finally, defining the decision variable $\mathbf{x} \in \mathbb{Z}^{\kappa^2-4\kappa +5}$ as the stack of all $z_{ij}$ and $\lambda$ in a column vector, the set of constraints can be written as $A_{\text{ineq}} \mathbf{x} \leq \mathbf{b}_{\text{ineq}}$, with proper $A_{\text{ineq}}$ and $\mathbf{b}_{\text{ineq}}$.

The first notable aspect is that we reformulated the problem in this manner at the expense of expanding the optimization space from $\kappa-2$ to $\kappa^2-4\kappa +5$ decision variables. Additionally, the objective function \eqref{fitnessfunction} remains nonlinear, categorizing the problem as a MINLP. However, the constraints have been formulated in a linear manner to enhance tractability. This decision is based on numerical experiments conducted on our specific problem, which revealed that solvers faced significant difficulties when handling nonlinear constraints in this context.

\subsubsection{Truncation-Based Integer Formulation}

When the number of available intermediate waypoints $\kappa -2$ is large, the permutation-based integer formulation becomes highly inefficient due to the dimension of the optimization space. We propose a truncation-based integer formulation as an alternative for the benchmarking. We define the decision variable $\mathbf{x}\in \mathbb{Z}^{\kappa -1}$ as the vector containing the waypoints labels $2 \leq \mathbf{x}_i \leq \kappa-1 \text{ for } i = 1, \dots, \kappa -2,$ and the number of intermediate waypoints to be held after truncation (excluding fixed first and last waypoints) $0 \leq \mathbf{x}_{\kappa-1} \leq \kappa-2$. Then, based on the baseline fitness function \eqref{fitnessfunction}, the objective function of this truncation-based integer formulation $ \zeta(\mathbf{x})$ is defined as
\begin{equation*}
\zeta(\mathbf{x}) = h(\eta(\mathbf{x})),
\end{equation*}
where $\eta(\mathbf{x})$ is constructed as follows:
1) Let $ k = \mathbf{x}_{\kappa-1} $, the last entry of $\mathbf{x} $; 2) Extract and stack into a vector the first $ k$-entries of $ \mathbf{x} $: $\mathbf{x}_{1:k} = (\mathbf{x}_1, \mathbf{x}_2, \dots, \mathbf{x}_k) $; 3) Define $\rho(\cdot)$ as the operator that removes duplicates from a vector, then $\eta(\mathbf{x}) = \rho(\mathbf{x}_{1:k})$.

In this way, the objective function $\zeta(\mathbf{x})$ \hl{evaluates} a duplicate-free \hl{variable-length} sequence using the baseline fitness function \eqref{fitnessfunction}. Although this method lacks the mathematical elegance of the permutation-based integer formulation, it provides a practical and straightforward solution for handling variable-length and unique waypoint assignment sequences without increasing the dimension of the optimization space or adding nonlinear constraints to handle, for instance, the duplicate-free requirement using a penalty method.

\subsubsection{Benchmarking Results}
Both the permutation-based and truncation-based integer formulations are expressed in a MINLP framework, enabling the use of solvers designed for such problems. Since our algorithm was developed in MATLAB, we compare it against solvers available in the same environment—specifically, MATLAB's Genetic Algorithm and Surrogate Optimization tools, the two methods currently capable of handling MINLP within MATLAB (R2024b). Therefore, we have four methods to compare against our algorithm, namely, the Permutation-based Integer Formulation solved \hl{by MATLAB's} GA (PIFGA), the Permutation-based Integer Formulation solved by the Surrogate Optimization (PIFSO), the Truncation-based Integer Formulation solved \hl{by MATLAB's} GA (TIFGA), and the Truncation-based Integer Formulation solved by the Surrogate Optimization (TIFSO).

\begin{figure}[!t]
\centering
\includegraphics[width=0.30\textwidth]{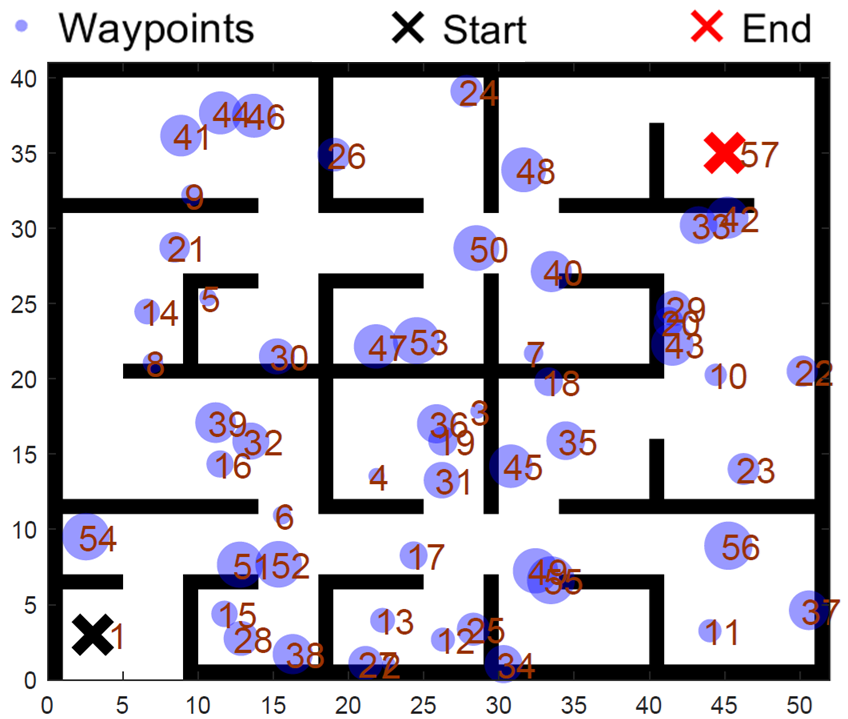}
\caption{Benchmarking map, example case with 55 intermediate waypoints, fixed start and final locations. A larger blue circle indicates a higher reward.}
\label{fig:scenario55}
\end{figure}

The benchmark environment comprises 80 scenarios, consisting of 20 randomly generated waypoint distributions for each of the four number of intermediate waypoints: 10, 25, 40, and 55, with fixed initial and final location. This setup allows us to evaluate our algorithm across varying levels of complexity. For reference, one of the scenarios with 55 intermediate waypoints is shown in Fig. \ref{fig:scenario55}. We use the same differential-drive vehicle model \eqref{differentialdynamics}, without time restriction, maximum distance of $200$ m, maximum speed of $1$ m/s, a minimum speed of $0.1$ m/s, and a maximum angular speed of $0.5$ rad/s.

\begin{figure}[!t]
\centering
\includegraphics[width=0.5\textwidth]{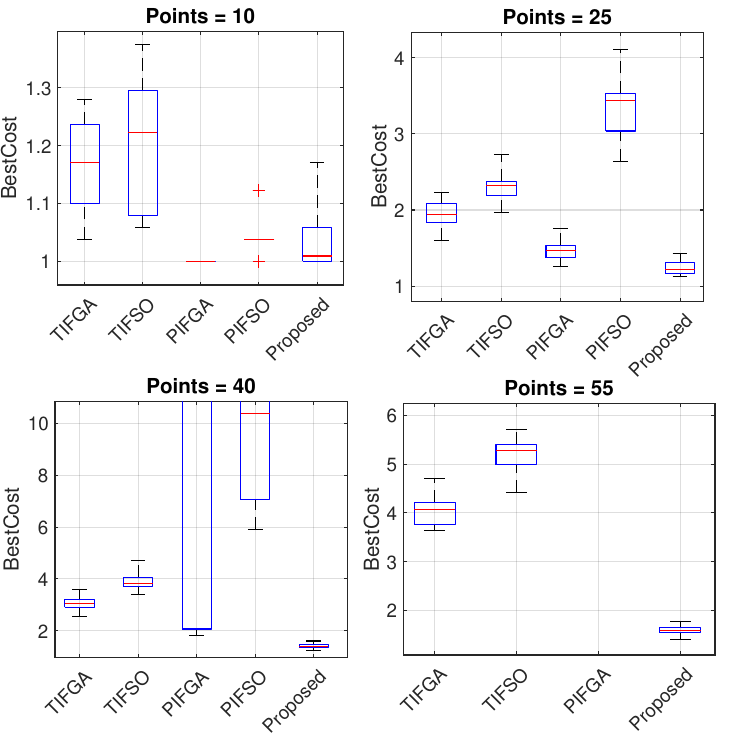}
\caption{Benchmarking results sorted by the number of intermediate waypoints. In the case of 55 waypoints, the PIFGA method is out of the plot as its \texttt{BestCost} \hl{exceeds} 5000, and the PIFSO method does not produce results due to its inability to handle this number of parameters. Very high values of the \texttt{BestCost} indicate that the penalty function is excessively large, indicating that no feasible solution was found.}
\label{fig:bench1}
\end{figure}

\begin{figure}[!t]
\centering
\includegraphics[width=0.35\textwidth]{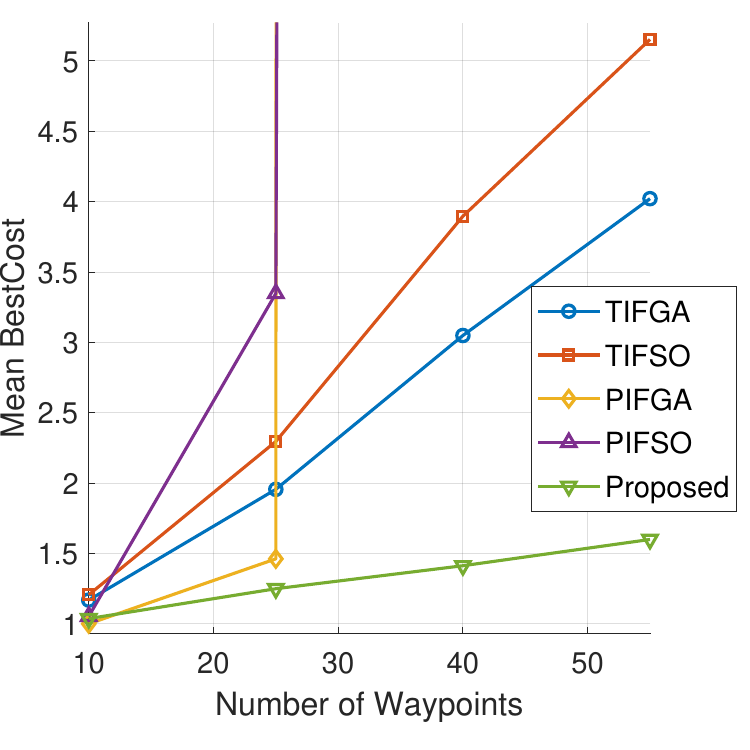}
\caption{The benchmarking results reported in terms of the mean best fitness function obtained (minimization problem). The proposed method consistently outperforms the alternatives, even as the number of points increases. The permutation-based formulations degrade rapidly as the number of waypoints increases.}
\label{fig:bench2}
\end{figure}

The results are presented as box-and-whisker diagrams in Fig. \ref{fig:bench1}, where the proposed method outperforms the four alternatives across all waypoint scenarios, except in the case of the smallest number of waypoints (10), where the PIFGA slightly more frequently achieves the best solution. It can be observed that the methods based on permutation perform well for the smallest number of waypoints, but their performance rapidly deteriorates as the number of waypoints increases, probably due to the rapid growth of the optimization space dimension $ \kappa^2 -4\kappa + 5 $. Conversely, the truncation-based algorithms maintain relatively fair results in scenarios with a large number of waypoints. Fig. \ref{fig:bench2} provides a clearer comparison of the scalability of each algorithm by showing how the mean best fitness function value varies with the number of intermediate waypoints.

\subsection{Time-Complexity Study}

\begin{figure}[!t]
\centering
\includegraphics[width=0.35\textwidth]{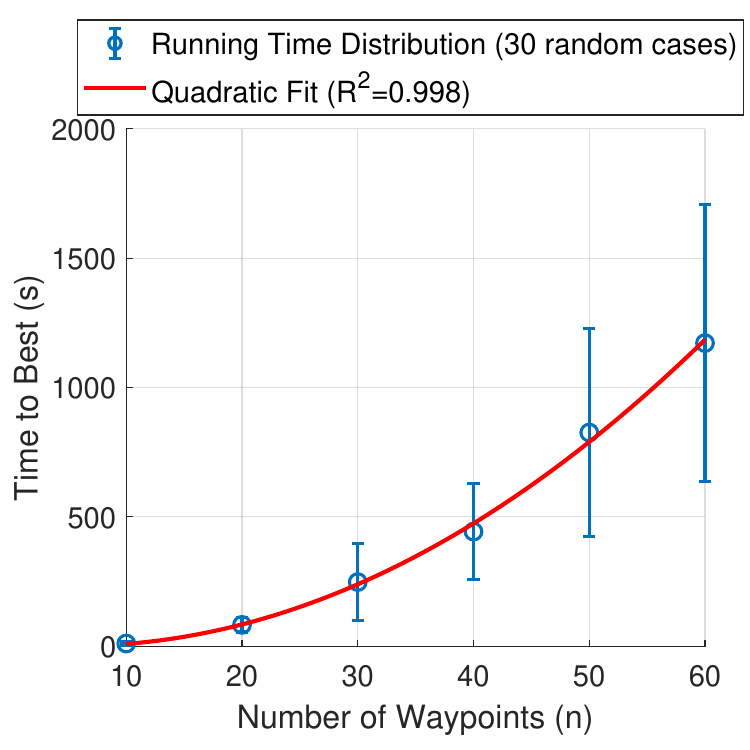}
\caption{The running time complexity of the proposed algorithm grows approximately as $\mathcal{O}(n^2)$, based on empirical analysis and quadratic fitting of the data. A total of 180 simulations with randomly distributed waypoints were conducted, consisting of 30 scenarios for each waypoint count: 10, 20, 30, 40, 50, and 60.}
\label{fig:timecomplex}
\end{figure}

To evaluate the time complexity of the proposed algorithm, we conducted 180 simulations across 30 randomly generated waypoint distributions for each of the following total waypoint counts: 10, 20, 30, 40, 50, and 60. The relationship between the time required to achieve the best-found solution and the number of intermediate waypoints is illustrated in Fig. \ref{fig:timecomplex}. Furthermore, as depicted in the same figure, the algorithm exhibits a time complexity that appears to scale approximately as $\mathcal{O}(n^2)$. This complexity aligns with the expected behavior since we used nested loops to calculate the fitness value for each candidate in each generation. Compared to this complexity, the brute force (guaranteed optimal solution) of the classical TSP alone (without dynamics or obstacles) is $\mathcal{O}(n!)$, and the best-known solution for TSP is the Held–Karp algorithm \cite{held1962dynamic} with time complexity being $\mathcal{O}(n^2 2^n)$.

\section{Limitations and Future Work}
\label{sec:futurework}

Although the proposed algorithm demonstrates strong performance and scalability across various robot platforms, several limitations remain, presenting avenues for future exploration. One key limitation is its current focus on single-robot mission planning. Extending this approach to multi-robot systems would enhance its applicability, particularly for collaborative missions \cite{robotics13030040}. Future work could address inter-robot communication, coordination, and conflict resolution to enable efficient multi-robot operations. Incorporating swarm intelligence or decentralized approaches may also be fruitful in this context.

Another area for improvement is the handling of dynamic obstacles. While the algorithm effectively navigates predefined, static obstacle environments, real-time adaptation to dynamic obstacles requires further development. Integrating predictive models or reactive obstacle avoidance techniques could enhance robustness in dynamic settings. \hl{As a future improvement, we propose that when the environment changes during mission execution, seeding the initial population with the offline-computed reference solution can enable the algorithm to rapidly converge to a new, feasible, albeit suboptimal, solution.}
Similarly, real-time constraint handling remains an open challenge. In the current framework, constraints are validated offline during the candidate solution evolution phase, producing a trajectory that maximizes total reward. However, once execution begins, the algorithm does not re-plan the trajectory, leaving the actual control strategy—whether predictive, reactive, or otherwise—independent of the offline planning process. While this flexibility supports a broad range of controllers, real-time corrections to handle unexpected violations warrant further investigation.

Beyond waypoint-based formulations, generalizing the algorithm to incorporate task-specific time and energy costs, as well as modeling interdependencies between tasks, would expand its applicability. Future work could also consider tasks that carry not only a reward but also an associated risk, particularly in heterogeneous multi-agent systems \cite{PRASAD2022110391}. Additionally, while the current formulation leverages differential flatness for efficient trajectory planning, many control systems are not inherently differentially flat. Developing methods to approximate or adapt to non-differentially flat systems would significantly broaden the range of potential applications, particularly in systems with complex dynamics or constraints.

To achieve fully autonomous robot operation, with human intervention only when necessary, a promising future direction involves inferring human intentions from demonstrations using data-driven techniques such as inverse optimal control \cite{10383220} or inverse reinforcement learning \cite{liang2024online}. Once the intention is inferred, the robot could autonomously adjust its decision-making to align with the human's intent \cite{9849419,9852712}, even in \hl{real time} \cite{liang2024online}, enabling more responsive and adaptive interactions.

To ensure feasibility in real-time execution, further optimization is necessary, especially given the algorithm’s $\mathcal{O}(n^2)$ time complexity when dealing with a large number of waypoints. Implementing the algorithm in a faster programming language could enhance computational efficiency. We have already incorporated a parallel computing scheme, enabling concurrent fitness evaluations to reduce runtime. However, this improvement comes at the cost of increased memory usage, highlighting the trade-off between speed and resource consumption. Future efforts should explore strategies to balance these computational trade-offs while maintaining the algorithm’s scalability.

\section{Conclusions}
\label{sec:conclusions}

This paper introduces a new reward-based algorithm for mission planning of autonomous robots, extending the classical reward-based traveling salesman problem. Our approach integrates critical real-world constraints, including obstacle avoidance, differential dynamics, state and input limitations, mission time windows, and maximum travel distances. Additionally, it allows for the strategic bypassing of certain waypoints to satisfy constraints while maximizing the total reward. The problem is structured into two layers: the waypoint sequence is optimized using the proposed genetic algorithm, while motion planning is efficiently handled through clothoid curves and differential flatness. This approach avoids the computational burden of methods like MPC while achieving smoother trajectories than classical minimum jerk and snap-based techniques. Within the genetic algorithm, we propose a new crossover technique for sequences of different lengths in a discrete space. Our method, based on dynamic time warping, extended convex combination, and projection, provides a versatile tool applicable to other discrete optimization problems involving variable-length solutions.

We validated the proposed algorithm through diverse simulation scenarios, demonstrating adaptability to changes in waypoint rewards, mission time constraints, and newly introduced waypoints in unexplored regions. Experimental results with ground vehicles, quadrotors, and quadrupeds further confirm its effectiveness and versatility. Additionally, we present a simplified approach to handling quadruped dynamics using differential flatness from a kinematic control perspective, successfully implemented in hardware experiments.

To evaluate algorithmic efficiency, we reformulated the problem into two distinct MINLP approaches: one based on permutation and the other on truncation. We integrated MATLAB’s (R2024b) surrogate optimization and genetic algorithm into these formulations and benchmarked them against the proposed algorithm. The results demonstrate that the proposed algorithm outperforms nearly all scenarios and solvers, \hl{except} the permutation-based formulation solved by the genetic algorithm for the case of 10 waypoints. Furthermore, the proposed algorithm exhibited superior scalability, consistently achieving better solutions in scenarios with a high number of waypoints. Furthermore, the algorithm demonstrated a time complexity of approximately $\mathcal{O}(n^2)$.


%

\bibliographystyle{IEEEtran}

\bibliography{IEEEexample}{}

\newpage

\vspace{11pt}


\vfill

\end{document}